\documentclass{article} % For LaTeX2e
\usepackage{iclr2016_conference}
\usepackage{times}
\usepackage{wrapfig}
\usepackage{epsfig}
\usepackage{graphicx}
\usepackage{amsmath}
\usepackage{amssymb}
\usepackage{subcaption} 
\usepackage{hyperref}
\usepackage{url}

\title{What Happened to My Dog in That Network: \\ Unraveling Top-down Generators \\ in Convolutional Neural Networks}

\author{Patrick W. Gallagher \& Shuai Tang \& Zhuowen Tu \\
Department of Cognitive Science\\
University of California, San Diego\\
\texttt{\{pwgallag,shuaitang93,ztu\}@ucsd.edu} \\}

\begin{document}
\maketitle
\vspace{-4mm}
\begin{abstract}
\vspace{-3mm}
Top-down information plays a central role in human perception, but
plays relatively little role in many current state-of-the-art
deep networks, such as Convolutional Neural Networks (CNNs). This work seeks to 
explore a path by which top-down information can have a direct impact within current 
deep networks. We explore this path by learning and using ``generators'' corresponding 
to the network internal effects of three types of transformation (each a restriction
of a general affine transformation): rotation, scaling, and translation.
We demonstrate how these learned generators can be used to transfer top-down information
to novel settings, as mediated by the ``feature flows'' that the transformations 
(and the associated generators) correspond to inside the network. Specifically, we explore 
three aspects: 1) using generators as part of a method for synthesizing transformed images 
--- given a previously unseen image, produce versions of that image corresponding to 
one or more specified transformations; 2) ``zero-shot learning'' --- when provided 
with a feature flow corresponding to the effect of a transformation of unknown amount, 
leverage learned generators as part of a method by which to perform an accurate categorization 
of the amount of transformation, even for amounts never observed during training; 
and 3) (inside-CNN) ``data augmentation'' --- improve the classification performance 
of an existing network by using the learned generators to directly provide 
additional training ``inside the CNN''. 
\end{abstract}

\vspace{-4mm}
\section{Introduction}
\vspace{-3mm}
Deep learning has many recent successes; for example, deep learning
approaches have made strides in automatic speech recognition \citep{hinton2012deep},
in visual object recognition \citep{krizhevsky2012imagenet}, and in
machine translation \citep{sutskever2014sequence}. While these successes
demonstrate the wide-ranging effectiveness of deep learning approaches,
there yet remains useful information that current deep learning is less
able to bring to bear.

\vspace{-1mm}
To take a specific example, consider that much of current deep learning
practice is dominated by approaches that proceed from input to output
in a fundamentally bottom-up fashion. While current performance is extremely 
impressive, these strongly bottom-up characteristics leave room for one to ask 
whether providing deep learning with the ability to also incorporate top-down 
information might open a path to even better performance.

\vspace{-1mm}
The demonstrated role of top-down information in human perception \citep{stroop1935studies,cherry1953some,hill2007hollow,ames1951visual}
provides a suggestive indication of the role that 
top-down information could play in deep learning. Visual illusions
(such as the ``Chaplin mask'') provide the clearest examples of the strong
effect that top-down/prior information can have on human perception;
the benefits of top-down information in human perception
are widespread but subtler to notice: prominent examples include color
constancy \citep{kaiser1996human} and the interpretation of visual
scenes that would otherwise be relatively meaningless (e.g. the ``Dalmatian'' image 
\citep{marr1982vision}). Another particularly common experience is the human ability 
to focus on some specific conversation in a noisy room, distinguishing the relevant 
audio component among potentially overwhelming interference.

\vspace{-1mm}
Motivated by the importance of top-down information in human perception, as well
as by the successful incorporation of top-down information in non-deep
approaches to computer vision \citep{borenstein2008combined,tu2005image,levin2009learning},
we pursue an approach to bringing top-down information into current deep
network practice. The potential benefits from incorporating top-down
information in deep networks include improved prediction accuracy
in settings where bottom-up information is misleading or insufficiently
distinctive as well as generally improved agreement when multiple classification 
predictions are made in a single image (such as in images containing multiple objects).
A particularly appealing direction for future work is the use of top-down information 
to improve resistance to ``adversarial examples'' \citep{nguyen2015deep,szegedy2013intriguing}.

\vspace{-2mm}
\subsection{Related work}
\vspace{-2mm}
The incorporation of top-down information in visual tasks stands at
the intersection of three fields: cognitive science, computer vision,
and deep learning. Succinctly, we find our inspiration in cognitive
science, our prior examples in computer vision, and our actual instantiation
in deep learning. We consider these each in turn.

\vspace{-2mm}
\paragraph{Cognitive science}
\vspace{-2mm}
Even before Stroop's work \citep{stroop1935studies} it has been noted
that human perception of the world is not a simple direct path from,
e.g., photons reaching the retina to an interpretation of the world
around us. 
% Notwithstanding the claims of ``direct perception'' made by Gibson \citep{gibson1966senses,gibson2002theory}
Researchers have established a pervasive and important role for top-down information
in human perception \citep{gregory1970intelligent}. The most striking
demonstrations of the role of top-down information in human perception
come in the form of ``visual illusions'', such as incorrectly perceiving
the concave side of a plastic Chaplin mask to be convex \citep{hill2007hollow}. 

\vspace{-2mm}
The benefits of top-down information are easy to overlook, simply
because top-down information is often playing a role in the smooth
functioning of perception. To get a sense for these benefits, consider
that in the absence of top-down information, human perception would
have trouble with such useful abilities as the establishment of color
constancy across widely varying illumination conditions \citep{kaiser1996human}
or the interpretation of images that might otherwise resemble an unstructured
jumble of dots (e.g., the ``Dalmatian'' image \citep{marr1982vision}).

\vspace{-2mm}
\paragraph{Non-deep computer vision}
\vspace{-2mm}
Observations of the role of top-down information in human perception
have inspired many researchers in computer vision. A widely-cited
work on this topic that considers both human perception and machine
perception is \citep{kersten2004object}. The chain of research stretches
back even to the early days of computer vision research, but more
recent works demonstrating the performance benefits of top-down information
in tasks such as object perception include \citep{borenstein2008combined,tu2005image,levin2009learning}.

\vspace{-2mm}
\paragraph{Deep computer vision}
\vspace{-2mm}
Two recent related works in computer vision are \citep{cohen2015transform,jaderberg2015spatial}. 
There are distinct differences in goal and approach, however. Whereas spatial transformer 
networks \citep{jaderberg2015spatial} pursue an architectural addition in the form 
of what one might describe as ``learned standardizing preprocessing'' inside the 
network, our primary focus is on exploring the effects (within an existing CNN) 
of the types of transformations that we consider. We also investigate a method of 
using the explored effects (in the form of learned generators) to improve vanilla 
AlexNet performance on ImageNet. On the other hand, \citep{cohen2015transform} state 
that their goal is ``to directly impose good transformation properties of a representation 
space'' which they pursue via a group theoretic approach; this is in contrast to 
our approach centered on effects on representations in an existing CNN, namely AlexNet.  
They also point out that their approach is not suitable for dealing with images much 
larger than 108x108, while we are able to pursue an application involving the entire 
ImageNet dataset. Another recent work is \citep{dai2014generative}, modeling random 
fields in convolutional layers; however, they do not perform image 
synthesis, nor do they study explicit top-down transformations.
\vspace{-2mm}
\paragraph{Image generation from CNNs}
\vspace{-2mm}
As part of our exploration, we make use of recent work on generating
images corresponding to internal activations of a CNN. A special purpose
(albeit highly intriguing) method is presented in \citep{dosovitskiy2015learning}.
The method of \citep{mahendran2014understanding} is generally applicable,
but the specific formulation of their inversion problem leads to generated
images that significantly differ from the images the network
was trained with. We find the technique of \citep{dosovitskiy2015inverting}
to be most suited to our purposes and use it in our subsequent visualizations.

\vspace{-2mm}
\paragraph{Feature flows}
\vspace{-2mm}
One of the intermediate steps of our process is the computation of
``feature flows'' --- vector fields computed using the SIFTFlow
approach \citep{liu2011sift}, but with CNN features used
in place of SIFT features. Some existing work has touched on the usefulness
of vector fields derived from ``feature flows''. A related but much more theoretical 
diffeomorphism-based perspective is \citep{joshi2000landmark}. Another early reference
touching on flows is \citep{simard1998transformation}; however, the
flows here are computed from image pixels rather than from CNN
features. \citep{taylor2010convolutional} uses feature flow fields
as a means of visualizing spatio-temporal features
learned by a convolutional gated RBM that is also tasked with an image
analogy problem. The ``image analogy'' problem is also present in the work \citep{memisevic2007unsupervised}
focusing on gated Boltzmann machines; here the image analogy is performed
by a ``field of gated experts'' and the flow-fields are again used
for visualizations. Rather than pursue a special purpose re-architecting
to enable the performance of such ``zero-shot learning''-type ``image
analogy'' tasks, we pursue an approach that works with an existing CNN trained 
for object classification: specifically, AlexNet \citep{krizhevsky2012imagenet}.

\vspace{-2mm}
\section{Generator learning}
\vspace{-2mm}
We will focus our experiments on a subset of affine image operations:
rotation, scaling, and translation. In order to avoid edge effects
that might arise when performing these operations on images where
the object is too near the boundary of the image, we use the provided
meta information to select suitable images. 

\vspace{-3mm}
\subsection{Pipeline for generator learning}
\label{sec:image_selection}
\vspace{-2mm}
We select from within the 1.3M images of the ILSVRC2014 CLS-LOC task
\citep{russakovsky2014imagenet}. In our experiments, we will rotate/scale/translate
the central object; we wish for the central object to remain entirely
in the image under all transformations. We will use bounding box information
to ensure that this will be the case: we select all images where the
bounding box is centered, more square than rectangular, and occupies
40-60\% of the pixels in the image.  We find that 91 images
satisfy these requirements; we will subsequently refer to these 91
images as our ``original images''.

\vspace{-3mm}
\subsubsection{Generating transformed image pairs}
\vspace{-2mm}
We use rotation as our running example. For ease of reference, we will
use the notation $I_{j}\left[\theta\right]$ to denote a transformed
version of original image $I_{j}$ in which the central object has
been rotated to an orientation of $\theta$ degrees; the original
image is $I_{j}\left[0^{\circ}\right]=I_{j}$. Using this notation, we can
consider image pairs in which the difference between one image and
the other is the amount of rotation of the central object. For example,
in the pair $\left(I_{j}\left[\theta_{\text{init}}\right],I_{j}\left[\theta_{\text{init}}+\Delta\theta\right]\right)$,
the central object is at orientation $\theta_{\text{init}}$ in the
first image and at orientation $\theta_{\text{init}}+\Delta\theta$
in the second. 

\begin{figure*}[!htp]
\vspace{-2mm}
\begin{center}
\begin{tabular}{c}
\includegraphics[height=0.45\linewidth]{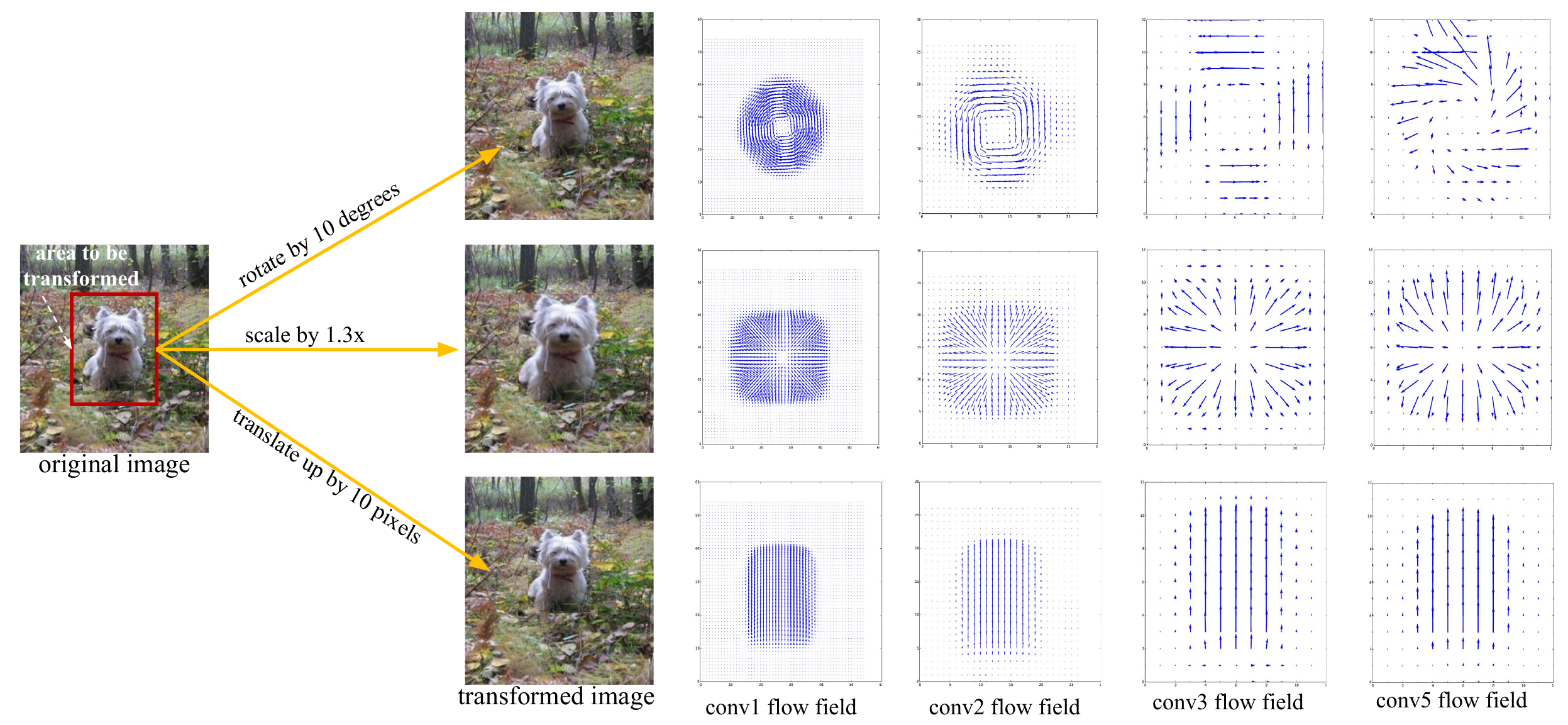}
\end{tabular}
\end{center}
\vspace{-6mm}
\caption{\footnotesize Illustration of AlexNet feature flow fields associated with the specified transformations. Best viewed in color, on-screen.}
\label{fig:transformations_and_feature_flows}
\vspace{-2mm}
\end{figure*}

To begin, we will consider 72 regularly-spaced
values of the initial orientation angle, $\theta_{\text{init}}\in\left\{ 0^{\circ},5^{\circ},10^{\circ},\ldots,355^{\circ}\right\} ,$
but only one value of rotation amount, $\Delta\theta=10^{\circ}.$
This means that for each of the 91 original images, we will have 72
pairs of the form $\left(I_{j}\left[\theta_{\text{init}}\right],I_{j}\left[\theta_{\text{init}}+\Delta\theta\right]\right)$.
These 6,552 total pairs will be the focus of our subsequent processing
in the rotation experiments.

\vspace{-3mm}
\subsubsection{Computing AlexNet features }
\vspace{-2mm}
Next, for each transformed image pair, we use the Caffe \citep{jia2014caffe}
library's pretrained reference AlexNet model to compute the AlexNet
features associated with each image in the pair. Using the notation
$F_{j}\left[\theta\right]$ to denote the collection of all AlexNet
feature values resulting when image $I_{j}\left[\theta\right]$ is
the input, this means that we now have 6,552 ``collected AlexNet
features'' pairs of the form $\left(F_{j}\left[\theta_{\text{init}}\right],F_{j}\left[\theta_{\text{init}}+\Delta\theta\right]\right)$.
AlexNet has 8 layers with parameters: the first 5 of these are convolutional
layers (henceforth referred to as conv1, conv2, $\ldots$, conv5); the
final 3 are fully connected layers (henceforth referred to as fc6,
fc7, fc8). Our attention will be focused on the convolutional layers
rather than than fully connected layers, since the convolutional layers
retain ``spatial layout'' that corresponds to the original image
while the fully connected layers lack any such spatial layout. 

\vspace{-3mm}
\subsubsection{Computing per-layer feature flows}
\vspace{-2mm}
For ease of reference, we introduce the notation $F_{j,\ell}\left[\theta_{\text{init}}\right]$
to refer to the AlexNet features at layer $\ell$ when the input
image is $I_{j}\left[\theta_{\text{init}}\right].$ From the ``entire
network image features'' pair $\left(F_{j}\left[\theta_{\text{init}}\right],F_{j}\left[\theta_{\text{init}}+\Delta\theta\right]\right)$,
we focus attention on one layer at a time; for layer $\ell$ the relevant
pair is then $\left(F_{j,\ell}\left[\theta_{\text{init}}\right],F_{j,\ell}\left[\theta_{\text{init}}+\Delta\theta\right]\right).$
In particular, at each convolutional layer, for each such $\left(F_{j,\ell}\left[\theta_{\text{init}}\right],F_{j,\ell}\left[\theta_{\text{init}}+\Delta\theta\right]\right)$
pair we will compute the ``feature flow'' vector field that best
describes the ``flow'' from the values $F_{j,\ell}\left[\theta_{\text{init}}\right]$
to the values $F_{j,\ell}\left[\theta_{\text{init}}+\Delta\theta\right]$. 

We compute these feature flow vector fields using the SIFTFlow method
\citep{liu2011sift} --- however, instead of computing ``flow of SIFT
features'', we compute ``flow of AlexNet features''. See Figure
\ref{fig:transformations_and_feature_flows} for an illustration of these computed feature flow vector fields.
For a layer $\ell$ feature pair $\left(F_{j,\ell}\left[\theta_{\text{init}}\right],F_{j,\ell}\left[\theta_{\text{init}}+\Delta\theta\right]\right),$
we refer to the corresponding feature flow as $V_{j,\ell}\left[\theta_{\text{init}},\theta_{\text{init}}+\Delta\theta\right]$.
Recalling that we only compute feature flows for convolutional layers,
collecting the feature flow vector fields $V_{j,\ell}\left[\theta_{\text{init}},\theta_{\text{init}}+\Delta\theta\right]$
for $\ell\in\left\{ \text{conv1},\ldots,\text{ conv5}\right\} $ results
in a total\footnote{$8,522=2\times55^{2}+2\times27^{2}+3\times\left(2\times13^{2}\right)$} of $8,522$
values; we collectively refer to the collected-across-conv-layers
feature flow vector fields as $V_{j,:}\left[\theta_{\text{init}},\theta_{\text{init}}+\Delta\theta\right]$.
If we flatten/vectorize these feature flow vector field collections
for each pair and then row-stack these vectorized flow fields, we
obtain a matrix with 6,552 rows (one row per image pair) and 8,522
columns (one column per feature flow component value in conv1 through
conv5).

\vspace{-2mm}
\subsubsection{Feature flow PCA}
\vspace{-2mm}
In order to characterize the primary variations in the collected feature
flow vector fields, we perform PCA on this matrix of 6,552 rows/examples 
and 8,522 columns/feature flow component values. We retain the first
10 eigenvectors/principal component directions (as ``columns'');
each of these contains 8,522 feature flow vector field component values.

%\begin{wrapfigure}{r}{0.64\textwidth}
\begin{figure}[!htp]
\vspace{-0mm}
\begin{center}

\begin{tabular}{ccccc}
\hspace{-2mm} {\scriptsize mean} \hspace{-2mm} &
\includegraphics[width=0.16\textwidth]{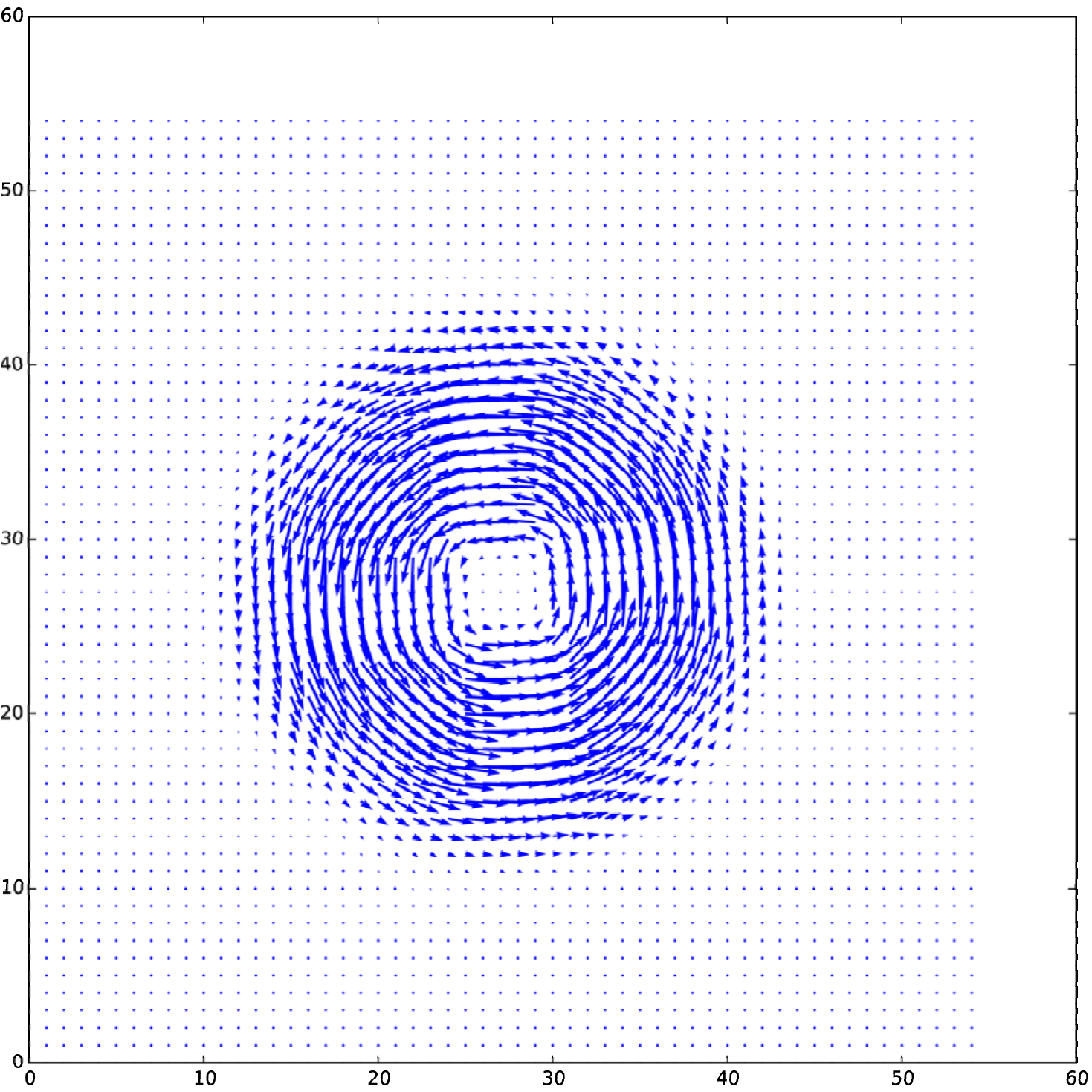} &
\hspace{-2mm} \includegraphics[width=0.16\textwidth]{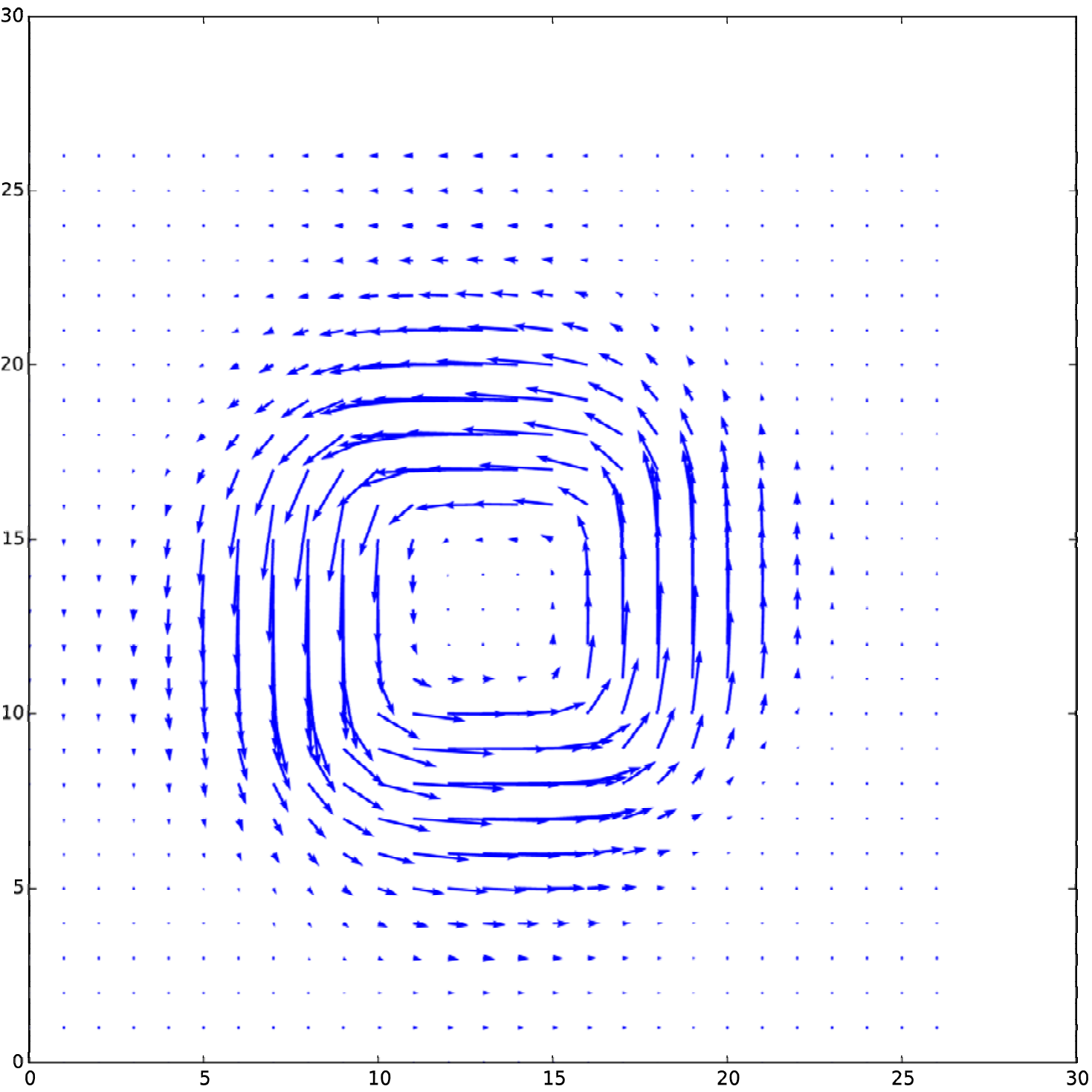} &
\hspace{-2mm} \includegraphics[width=0.16\textwidth]{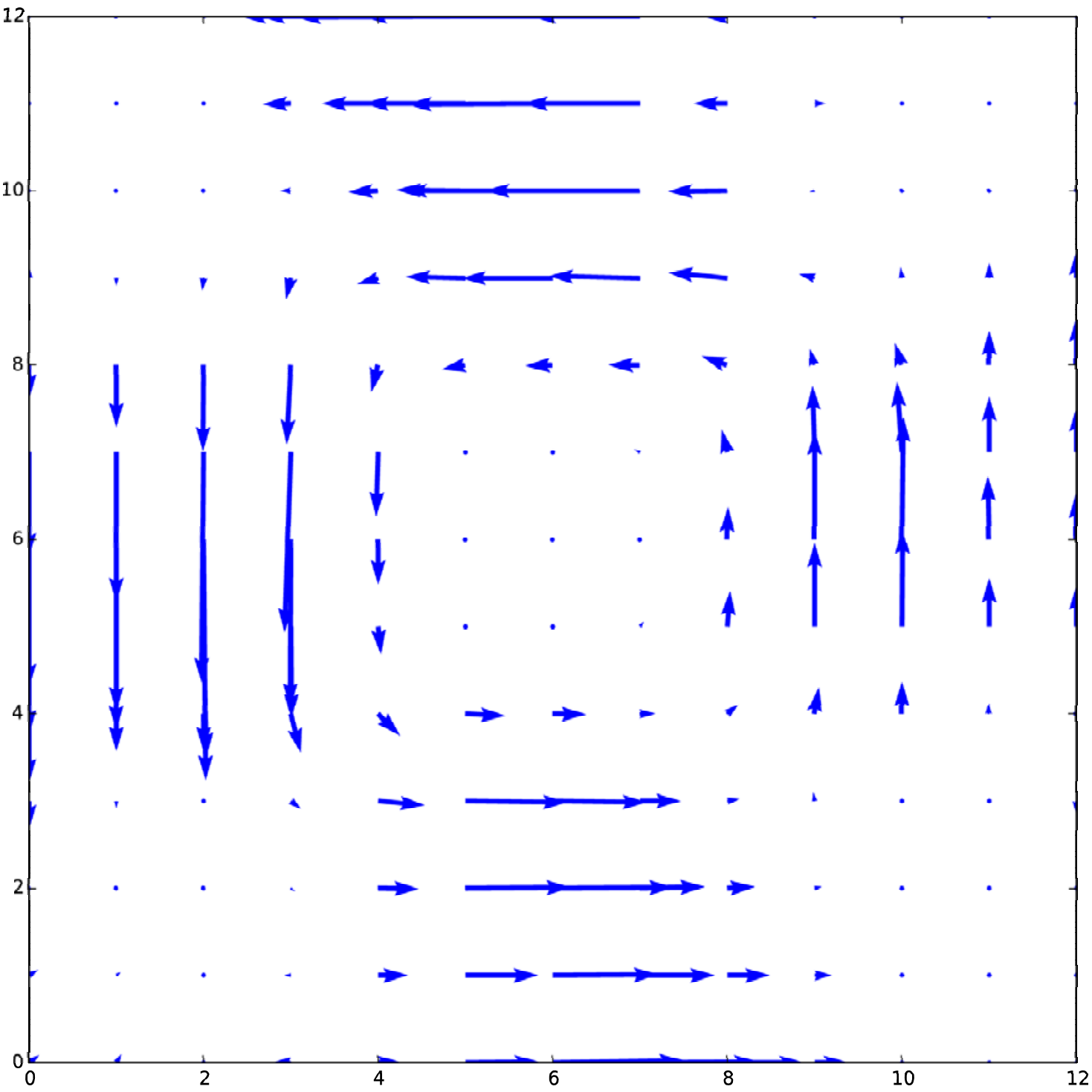} &
\hspace{-2mm} \includegraphics[width=0.16\textwidth]{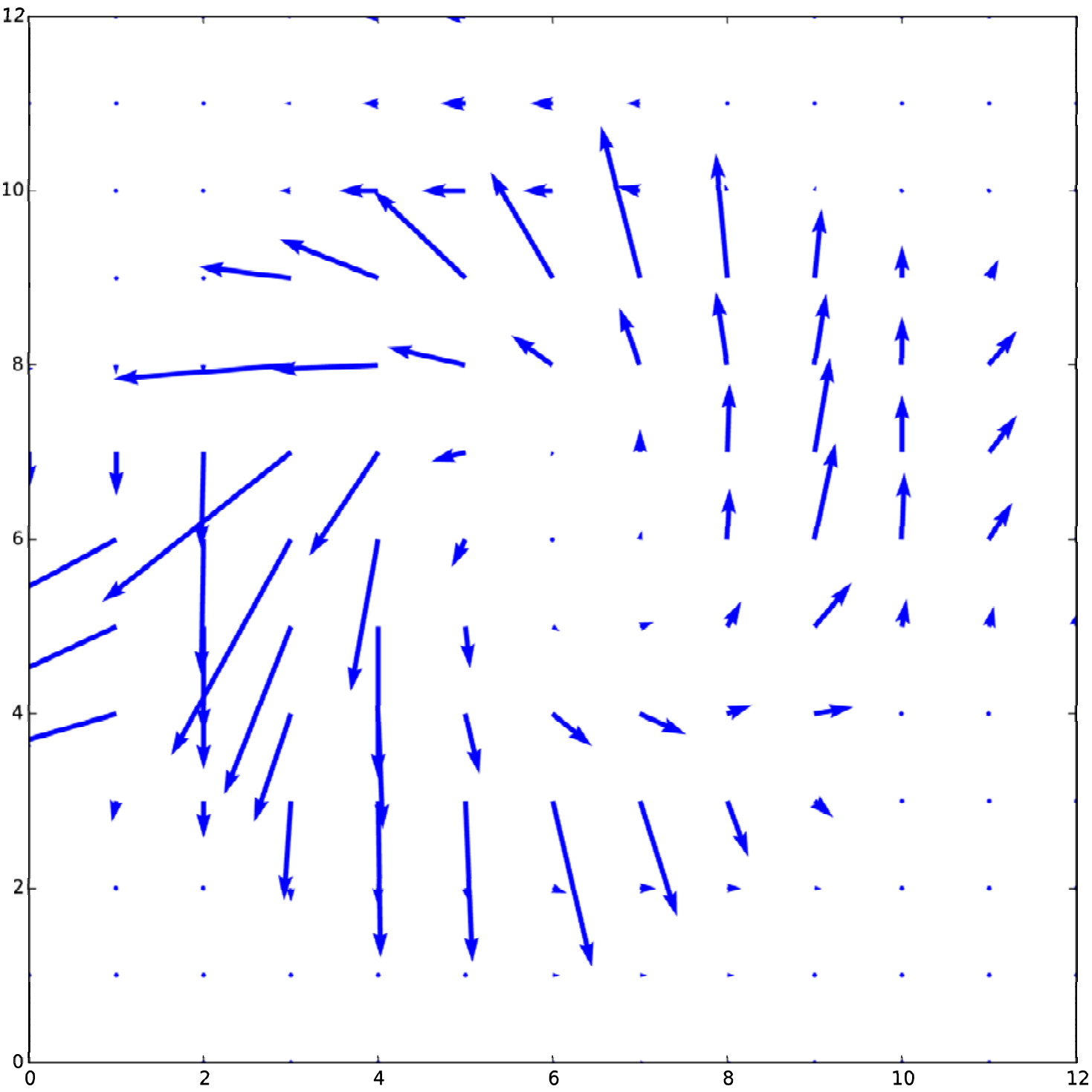} \\
\hspace{-2mm} {\scriptsize PC 1} \hspace{-2mm} & 
\includegraphics[width=0.16\textwidth]{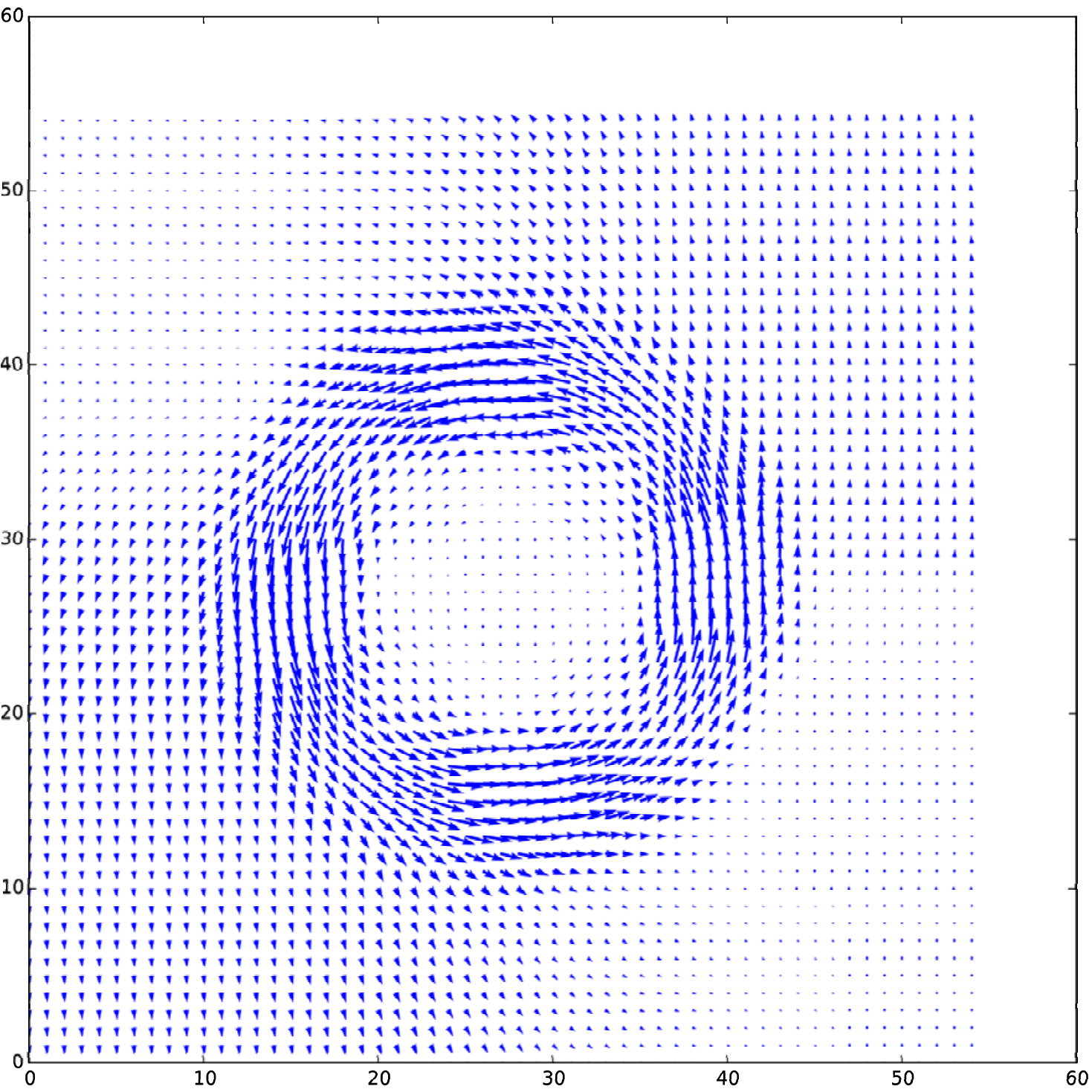} &
\hspace{-2mm}\includegraphics[width=0.16\textwidth]{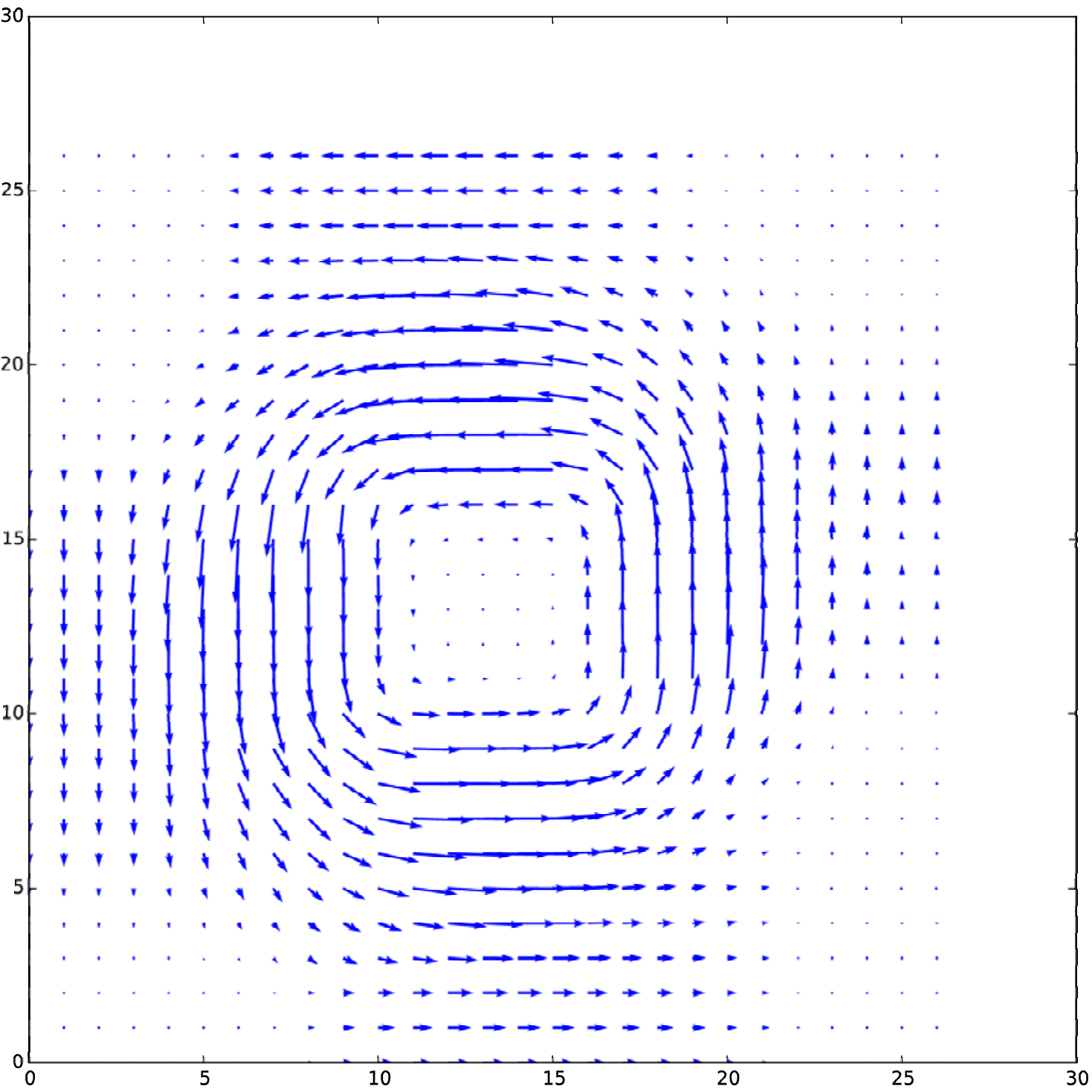} &
\hspace{-2mm} \includegraphics[width=0.16\textwidth]{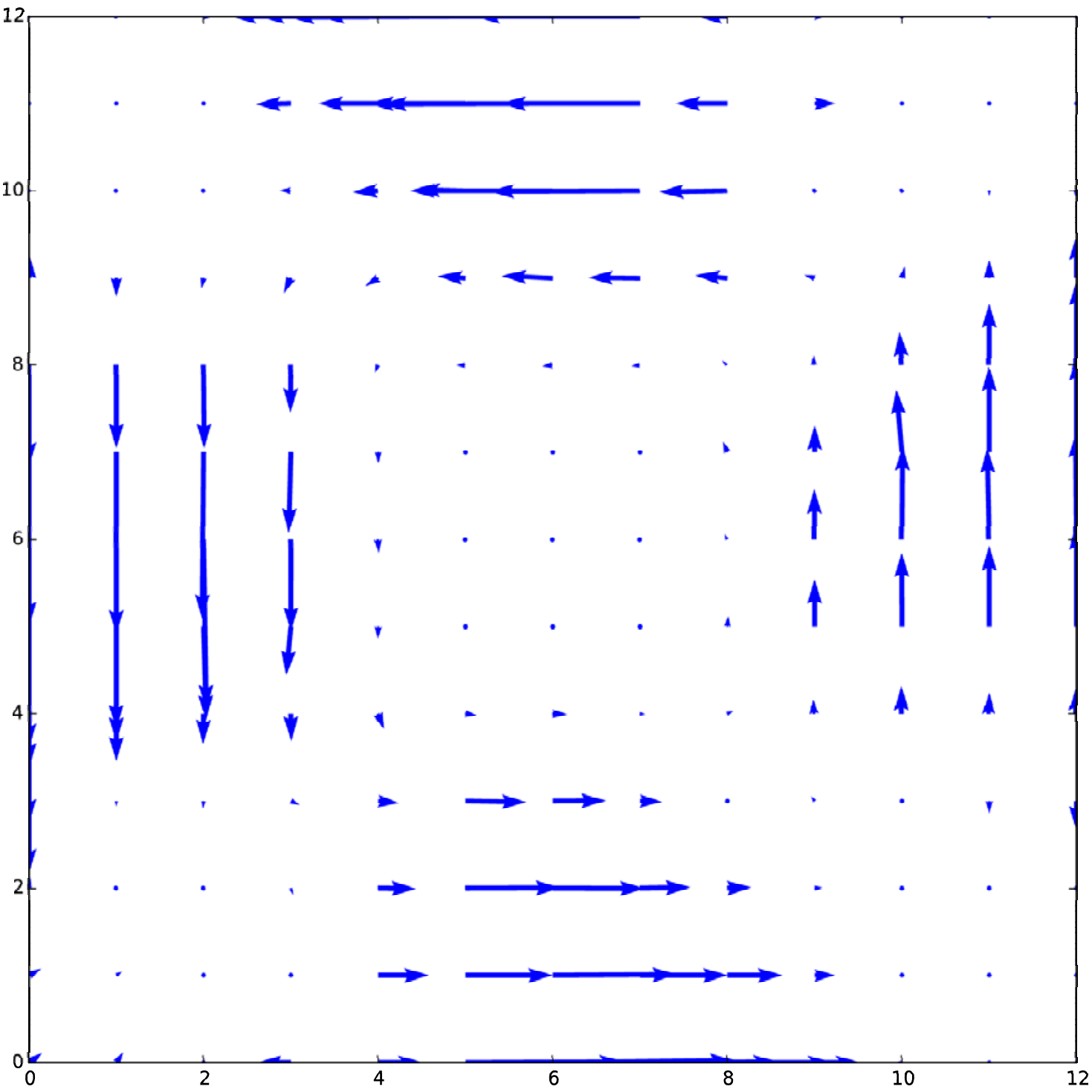} &
\hspace{-2mm} \includegraphics[width=0.16\textwidth]{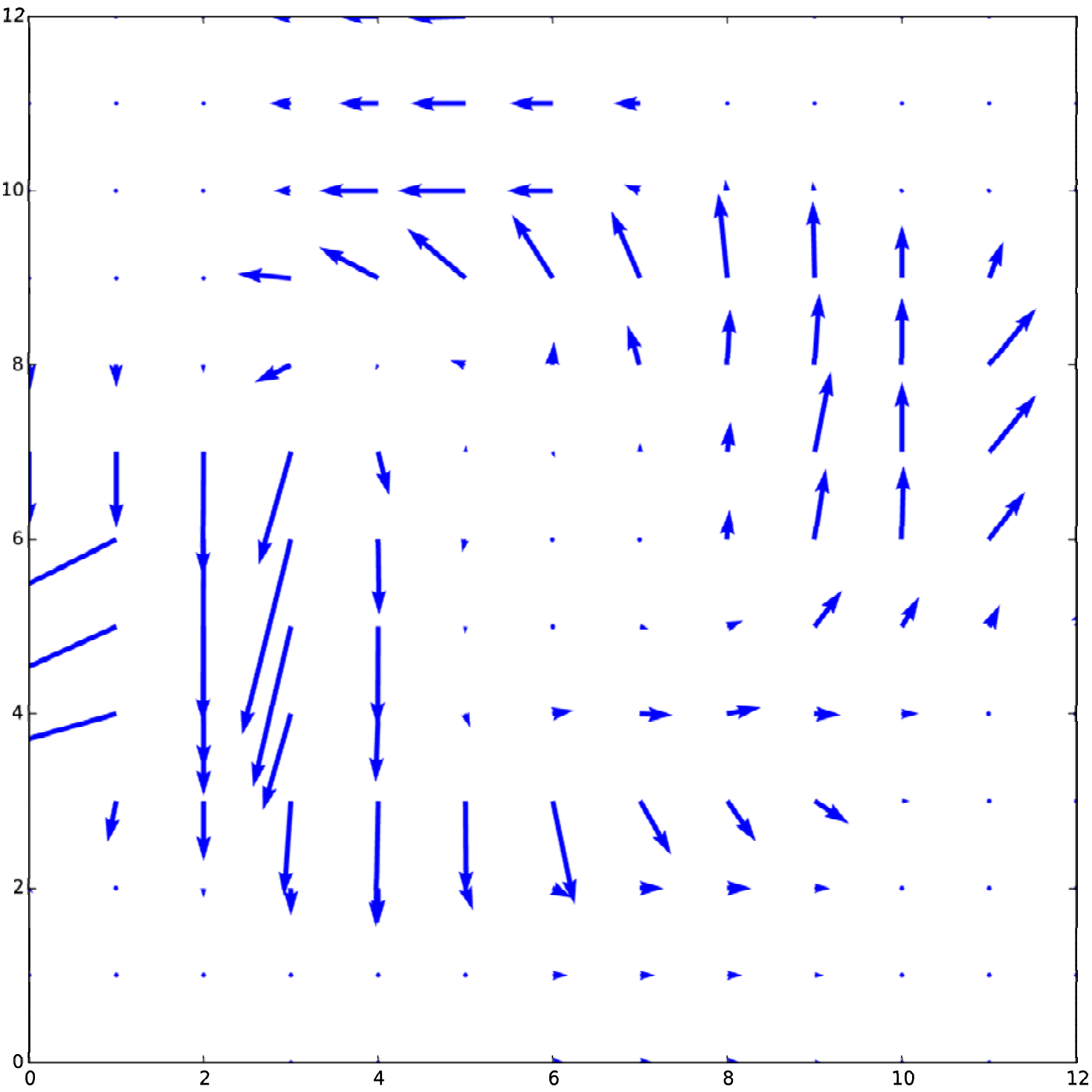} \\
\hspace{-2mm} {\scriptsize PC 2} \hspace{-2mm} & 
\includegraphics[width=0.16\textwidth]{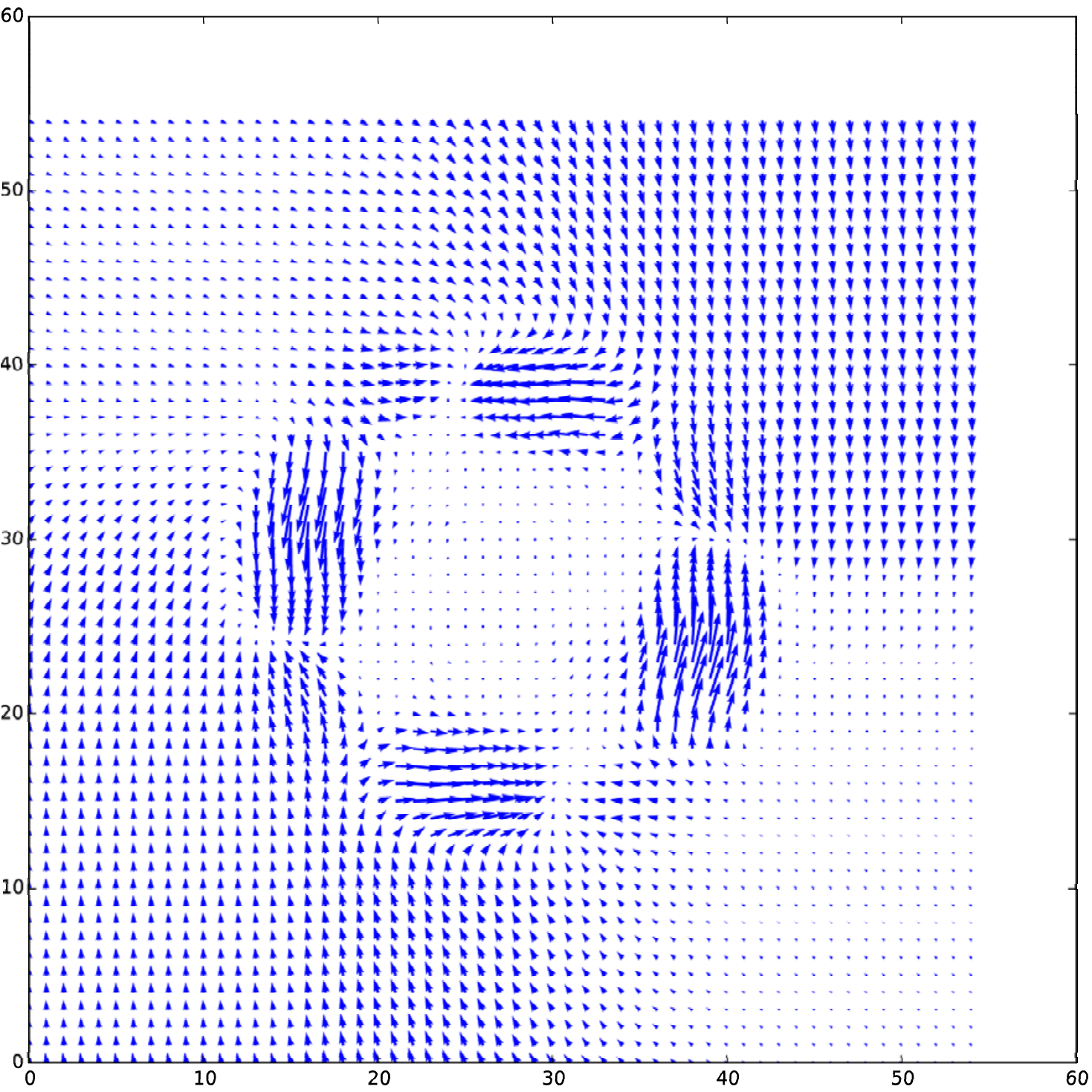} &
\hspace{-2mm}\includegraphics[width=0.16\textwidth]{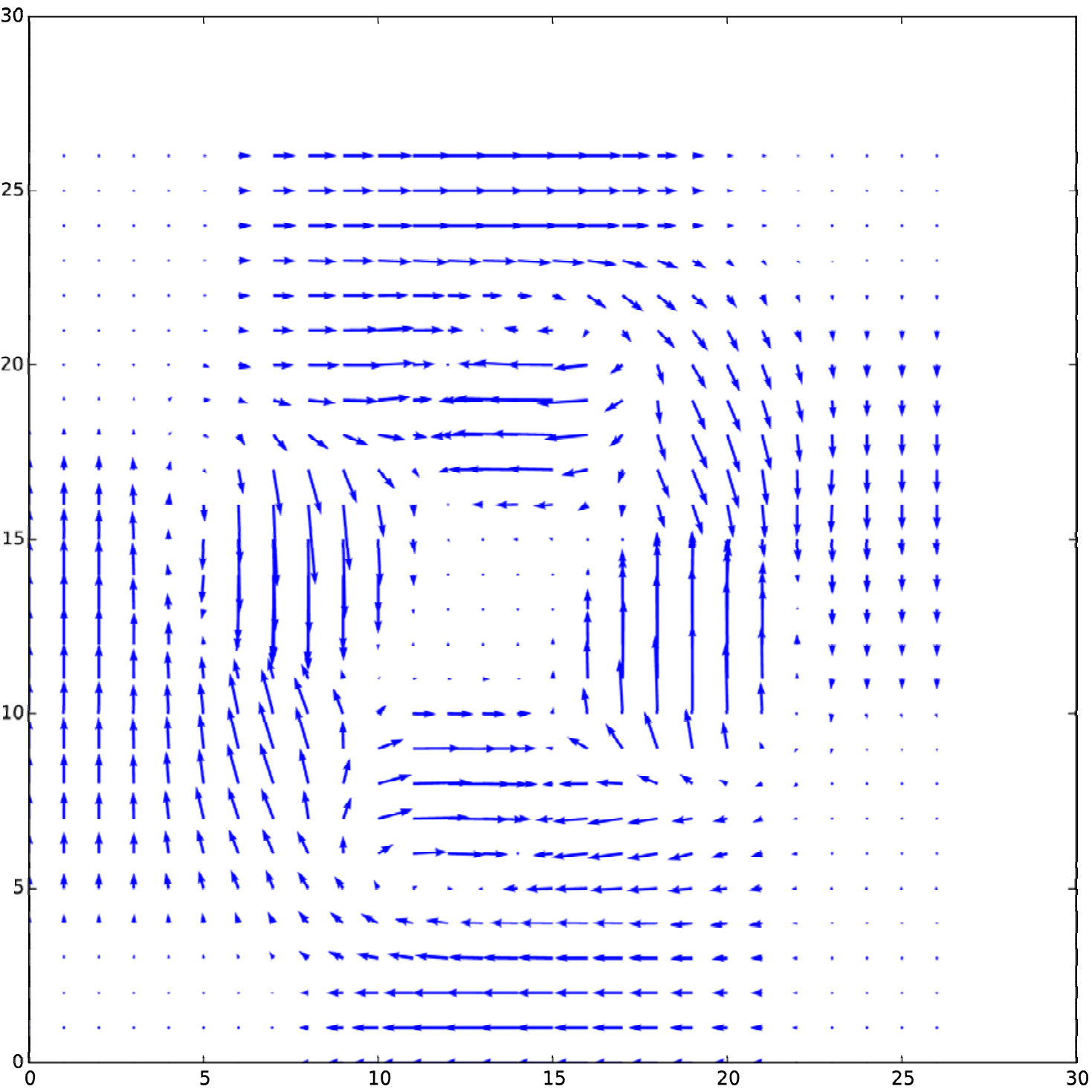} &
\hspace{-2mm}\includegraphics[width=0.16\textwidth]{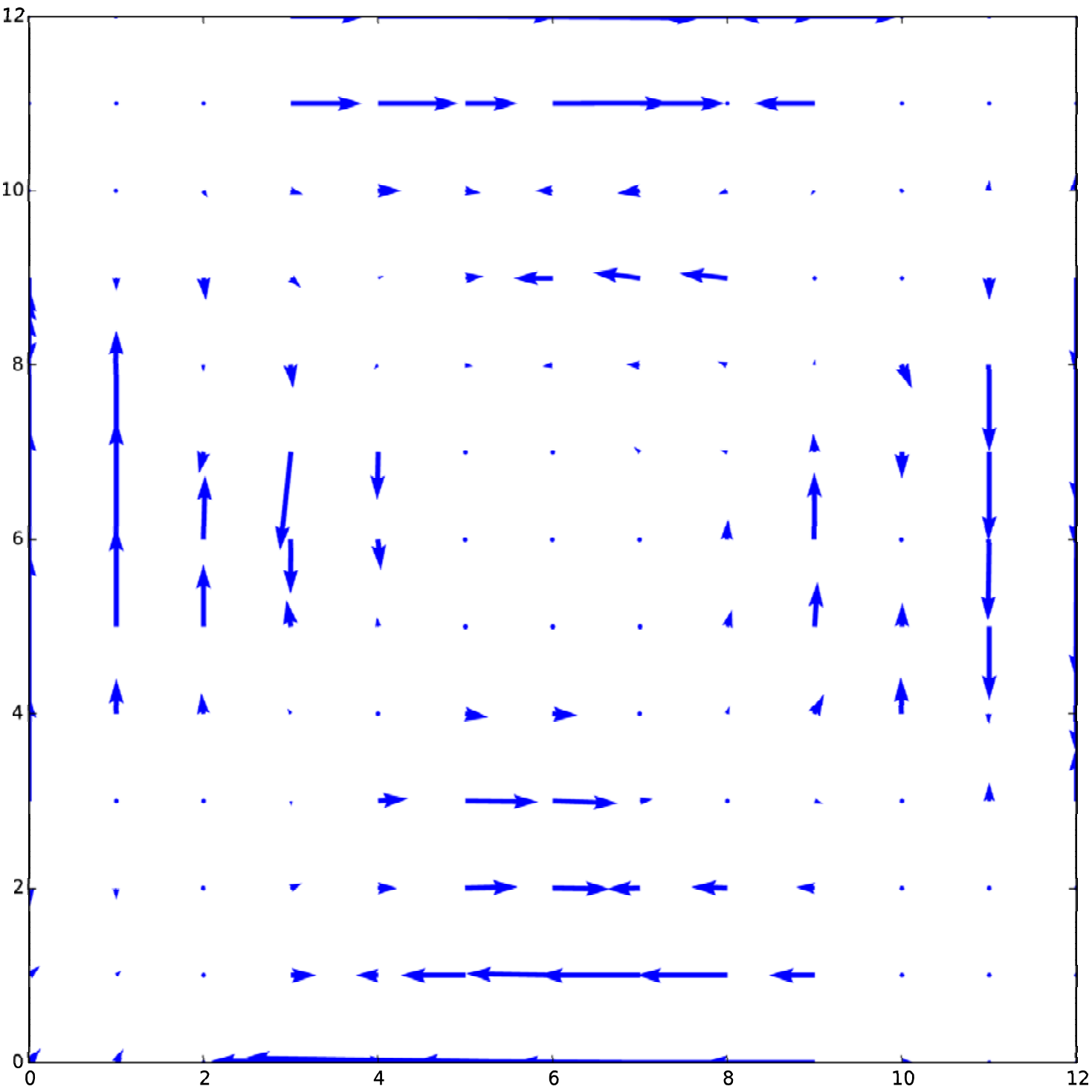} &
\hspace{-2mm} \includegraphics[width=0.16\textwidth]{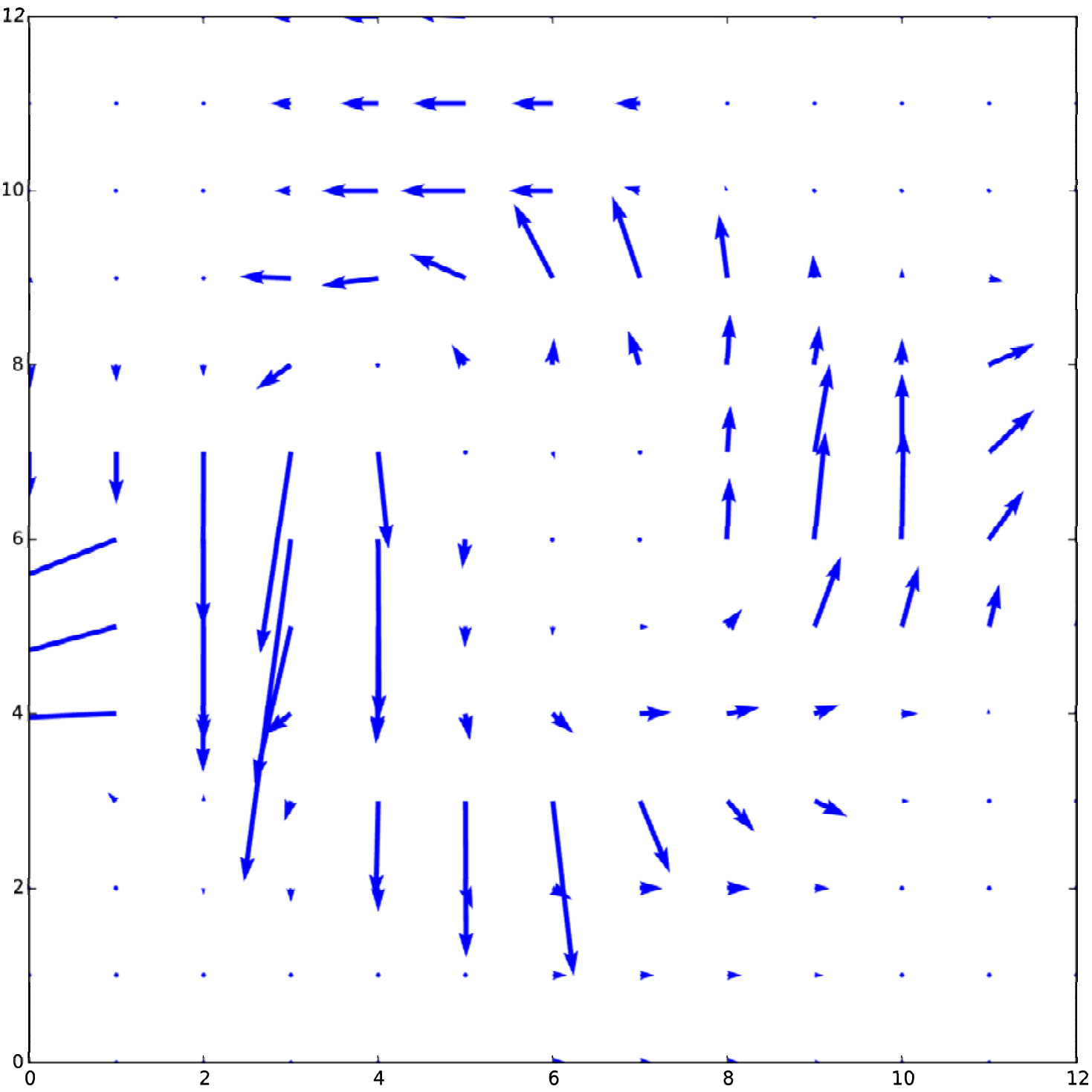} \\
\vspace{-1mm}
 & {\scriptsize conv1} & {\scriptsize conv2} & {\scriptsize conv3} & {\scriptsize conv5} \\
\end{tabular}\vspace{-3mm}
\caption{\footnotesize PCA components of the CNN feature flow fields associated with 10$^{\circ}$ of rotation. 
The first, second, and third rows show, respectively, the mean, first, and second principal components. 
The first, second, third, and forth columns show, respectively, the results in conv1, conv2, conv3, and conv5. Best viewed on screen.}
\label{fig:flow_pca}
\vspace{-2mm} 
\end{center}
%\end{wrapfigure}
\end{figure}

We denote these ``eigen'' feature flow fields as $\left\{ U_{1},\ldots,U_{10}\right\} ,$
with each $U_{i}\in\mathbb{R}^{8,522}.$ Here the use of an upper
case letter is intended to recall that, after reshaping, we can plot
these flow fields in a spatial layout corresponding to that of the
associated AlexNet convolutional layers. We also recall that these
``eigen'' feature flow fields were computed based on feature pairs with
a $10^{\circ}$ rotation. Together with the mean feature flow field,
subsequently denoted $M\in\mathbb{R}^{8,522}$, these 10 ``stacked
and flattened/vectorized `eigen' feature flow fields'' (as ``columns'')
of 8,522 feature flow component values provide us the ability to re-represent
each of the 6,552 ``per-pair stacked and flattened/vectorized feature
flow fields'' in terms of 11 = 1+10 coefficients. %[[1 ``coefficient
% of $M,$ the mean stacked and flattened/vectorized pair feature flow
% fields'' + 10 ``coefficients of $\left\{ U_{1},\ldots,U_{10}\right\} ,$
% the stacked and flattened/vectorized `eigen' feature flow fields'']]
When we re-represent the 6,552 example ``stacked and flattened/vectorized
'pair' feature flow fields'', the coefficient associated with the
mean will always be equal to 1; however, we will shortly consider
a setting in which we will allow the coefficient associated with the
mean to take on values other than 1.

\vspace{-2mm}
\subsubsection{Using feature flow PCA to obtain bases for expressing generators}
\vspace{-2mm}
Taken together, the mean feature flow field $M$ and the `eigen' feature
flow fields $\left\{ U_{1},\ldots,U_{10}\right\} ,$ provide us with
the ability to produce (up to some minimized value of mean squared
error) ``re-representations'' of each of the 6,552 example ``stacked and
flattened/vectorized `$10{}^{\circ}$ rotation' feature flow fields''.

These 11 vectors were determined from the case of feature flow fields
associated with 10$^{\circ}$ rotations. We next seek to use these
11 vectors (together with closely related additions described shortly)
as bases in terms of which we seek to determine alternative representations
of flow fields associated with other amounts of rotation. In particular, we will 
seek to fit regression coefficients for the representation of flow fields associated 
with feature pairs derived when there has been rotation of varying amounts of the 
central object. Specifically, we will follow the steps of the feature flow computation process detailed
earlier in this Section, but now using $\Delta\theta\in\left\{ 10^{\circ},20^{\circ},30^{\circ},40^{\circ},50^{\circ},60^{\circ}\right\} $
together with the previous $\theta_{\text{init}}\in\left\{ 0,5,10,\ldots,355\right\} .$ 

The regression equation associated with each feature flow example
will be of the form 
\vspace{-.1mm}
\begin{align}
U\left[\Delta\theta\right]\cdot w & =V_{j,:}\left[\theta_{\text{init}},\theta_{\text{init}}+\Delta\theta\right],\label{eq:single-example-regress-feature-flows-from-10-degree-flow-basis}
\end{align}
\vspace{-.1mm}
where $U\left[\Delta\theta\right]\in\mathbb{R}^{8,522\times33}$ is
a matrix containing 11 groups of 3 columns, each of the form $U_{i},\ \left(\frac{\Delta\theta}{10}\right)U_{i},\ \left(\frac{\Delta\theta}{10}\right)^{2}U_{i};$
there is one such group for each of $\left\{ M,U_{1},\ldots,U_{10}\right\} .$
We do this so that the basis vectors provided in $U\left[\Delta\theta\right]$
will be different in different rotation conditions, enabling better
fits. The vector $w\in\mathbb{R}^{33}$ can similarly be regarded
as containing 11 groups of 3 coefficient values, say $w=\left(a_{M},b_{M},c_{M},\ldots,a_{U_{10}},b_{U_{10}},c_{U_{10}}\right)^{T}.$
Finally, the right hand side $V_{j,:}\left[\theta_{\text{init}},\theta_{\text{init}}+\Delta\theta\right]$
is an instance of the collected-across-conv-layers feature flow vector
fields described earlier. We have one of the above regression expressions
for each ``transform image'' pair $\left(I_{j}\left[\theta_{\text{init}}\right],I_{j}\left[\theta_{\text{init}}+\Delta\theta\right]\right)$;
since in our current setting have 91 original images, 72 initial orientation
angles $\theta_{\text{init}}\in\left\{ 0,5,10,\ldots,350,355\right\} $,
and 6 rotation amounts $\Delta\theta\in\left\{ 10,20,\ldots,60\right\} $,
we have a total of $39,312$ such pairs. For ease of reference, we
will refer to the vertically stacked basis matrices (each of the form
$U\left[\Delta\theta\right],$ with $\Delta\theta$ being the value
used in computing the associated ``transform image'' pair as in
the example regression described in Eqn. \ref{eq:single-example-regress-feature-flows-from-10-degree-flow-basis})
as $\mathbf{U}\in\mathbb{R}^{3.4e8\times33}.$ Similarly, we will
refer to the vertically stacked ``feature flow vector field'' vectors,
each of the form $V_{j,:}\left[\theta_{\text{init}},\theta_{\text{init}}+\Delta\theta\right]\in\mathbb{R}^{8,522}$
as $\mathbf{V}.$ 

Our ``modified basis'' regression problem\footnote{For specificity, we describe the row dimension of $\mathbf{U}$: A total of $3.4\times10^{8}$
rows that come from $39,312$ $=91\times72\times6$ vertically stacked
one-per-image-pair matrices, each with 8,522 rows. Thus, the total
number of rows is $3.4\times10^{8}=335,016,864=91$ original images
$\times$ 72 initial orientations $\times$ 6 rotation amounts $\times$
8,522 entries in the feature flow field collection per image pair.} is thus succinctly expressed as 
\vspace{-2mm}
\begin{align}
\underset{w\in\mathbb{R}^{33}}{\text{minimize }} & \frac{1}{2}\left\Vert \mathbf{U}w-\mathbf{V}\right\Vert _{2}^{2}.\label{eq:collectively-regress-feature-flows-from-10-degree-flow-basis}
\end{align}
\vspace{-3mm}
We will refer to the minimizing $w$ argument as $w^{\text{lsq}}\in\mathbb{R}^{33}.$

\vspace{-2mm}
\section{Illustration of use of learned generators}
\vspace{-2mm}
\subsection{Transformations via learned generators}
\vspace{-2mm}
We can use these ``least squares'' coefficient values $w^{\text{lsq}}\in\mathbb{R}^{33}$
to ``predict'' feature flow fields associated with a specified number
of degrees of rotation. More particularly, we can do this for rotation
degree amounts other than the $\Delta\theta\in\left\{ 10^{\circ},20^{\circ},30^{\circ},40^{\circ},50^{\circ},60^{\circ}\right\} $
degree amounts used when we generated the $39,312$ ``transform image
training pairs'' used in our least-squares regression calibration Eqn. \ref{eq:collectively-regress-feature-flows-from-10-degree-flow-basis}.
To obtain the desired generator, we decide what specific
``number of degrees of rotation'' is desired; using this specified
degree amount and the 11 basis vectors (learned in the 10 degree rotation
case we performed PCA on previously), generate the corresponding ``33
column effective basis matrix'' $U\left[\Delta\theta\right].$ Our
sought-for generator is then $U\left[\Delta\theta\right]\cdot w^{\text{lsq}},$
an element of $\mathbb{R}^{8,522}.$  For specificity, we will refer to the generator 
arising from a specified rotation angle of $\Delta\theta$ as $G\left[\Delta\theta\right]=U\left[\Delta\theta\right]\cdot w^{\text{lsq}}.$
We could describe  generators as ``predicted specified feature
flows''; however, since we use these to generate novel feature values in the layers of the network 
(and since this description is somewhat lengthy), we refer to them as ``generator 
flow fields'', or simply ``generators''.

% We now evaluate the quality of these learned generators by testing
% how closely each generator flow field matches the corresponding
% actual feature flow field (computed when the ``transform pair''
% involves the matching/specified amount of rotation).

\begin{figure}[!htp]
\vspace{-3mm}
\begin{centering}

% \begin{tabular}{cccccccc}
%  \includegraphics[width=0.11\textwidth]{figs/cat2_original_resized.png} &
% \hspace{-4mm} \includegraphics[width=0.11\textwidth]{figs/inverting_cat2_from_original_feature_conv1.png} & 
% \hspace{-4mm} \includegraphics[width=0.11\textwidth]{figs/inverting_cat2_from_warped_feature_conv1_rotated_30.png} & 
% \hspace{-4mm} \includegraphics[width=0.11\textwidth]{figs/inverting_cat2_from_warped_feature_conv1_rotated_-30.png} & 
% \hspace{-4mm} \includegraphics[width=0.11\textwidth]{figs/inverting_cat2_from_warped_feature_conv1_zoom_in_1_30.png} &
% \hspace{-4mm} \includegraphics[width=0.11\textwidth]{figs/inverting_cat2_from_warped_feature_conv1_zoom_out_1_30.png} &
% \hspace{-4mm} \includegraphics[width=0.11\textwidth]{figs/inverting_cat2_from_warped_feature_conv1_translate_-30_x_axis.png} & 
% \hspace{-4mm} \includegraphics[width=0.11\textwidth]{figs/inverting_cat2_from_warped_feature_conv1_translate_-30_y_axis.png} \\
% \hspace{-4mm} {\footnotesize (a) Input} & 
% \hspace{-4mm} {\footnotesize (b) ``Inverted''} & 
% \hspace{-4mm} {\footnotesize (c) $30^{\circ}$} &
% \hspace{-4mm} {\footnotesize (d) $-30^{\circ}$} &
% \hspace{-4mm} {\footnotesize (e) zoom $1.3$} &
% \hspace{-4mm} {\footnotesize (f) zoom $0.75$} &
% \hspace{-4mm} {\footnotesize (g) $-x,30$} &
% \hspace{-4mm} {\footnotesize (h) $-y,30$} 
% \end{tabular}\vspace{-3mm}
\includegraphics[width=.98\textwidth]{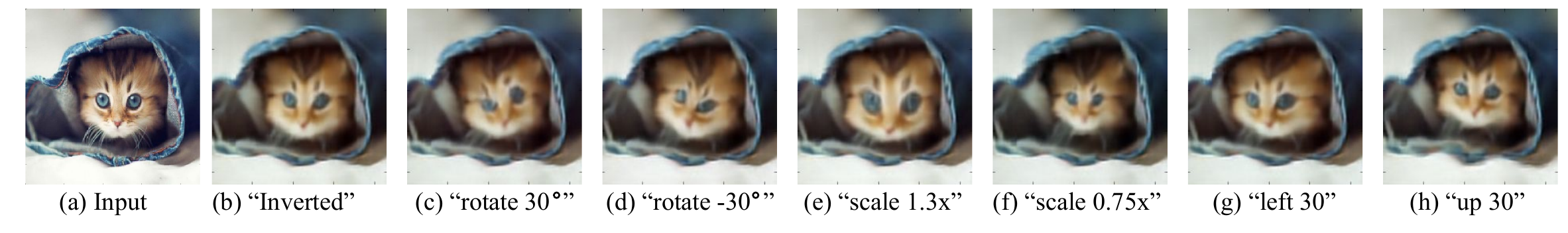}
\vspace{-3mm}
\caption{\footnotesize Illustration of applying learned generators to CNN features of a novel input image. 
(a) Input image. (b) ``Inverted image'' \citep{dosovitskiy2015inverting} from conv1 features of (a). 
(c), (d), (e), (f), (g), and (h) show, respectively, ``inverted images'' from conv1 features 
obtained by applying learned generator flow fields associated with -$30^{\circ}$ rotation, $30^{\circ}$ rotation,
scaling by a factor of $1.3$, scaling by a factor of $0.75$, translation $30$ pixels to the left, and translation $30$ pixels up.}
\label{fig:synthesis}
\vspace{-2mm} 
\par\end{centering}
\end{figure}

We may describe the use of these learned generators follows: given a collection of 
CNN features, we can apply a learned generator to obtain (approximations to) the feature 
values that would have arisen from applying the associated transformation to the input image.
% From a slightly different perspective, for any specific ``transform
% image'' pair $\left(I_{j}\left[\theta_{\text{init}}\right],I_{j}\left[\theta_{\text{init}}+\Delta\theta\right]\right)$
% we can produce a prediction (i.e. a generator), $U\left[\Delta\theta\right]\cdot w^{\text{lsq}}$,
% of the actual feature flow field $V_{j,:}\left[\theta_{\text{init}},\theta_{\text{init}}+\Delta\theta\right]\in\mathbb{R}^{8,522}$
% that would be produced by our pipeline steps.

We now seek to investigate the use of generator flow fields, generically $G\left[\Delta\theta\right]$,
in order to produce an approximation of the exact CNN features that would be observed if we were to
e.g. rotate an original image and compute the resulting AlexNet features. As a specific example, consider a ``transform
image pair'' $\left(I_{j}\left[\theta_{\text{init}}\right],I_{j}\left[\theta_{\text{init}}+\Delta\theta\right]\right)$.
In our notation, the corresponding AlexNet feature response map pair
is $\left(F_{j}\left[\theta_{\text{init}}\right],F_{j}\left[\theta_{\text{init}}+\Delta\theta\right]\right).$
We seek to use our generator, $G\left[\Delta\theta\right]\,$
to provide an estimate of $F_{j}\left[\theta_{\text{init}}+\Delta\theta\right]$
given only $F_{j}\left[\theta_{\text{init}}\right].$ 

Visualizations of the CNN features are often difficult to interpret. To provide an
interpretable evaluation of the quality of the learned generators, we use the 
AlexNet inversion technique of \citep{dosovitskiy2015inverting}. Applying our learned 
generators, the results in Fig. \ref{fig:synthesis} indicate that the resulting 
CNN features (arrived at using information learned in a top-down fashion) closely 
correspond to those that would have been produced via the usual bottom-up process. 
As a more quantitative evaluation, we also check the RMS error and mean absolute deviation
between network internal layer features ``generated'' using our learned generators 
and the corresponding feature values that would have arisen through ``exact'' bottom-up processing.
For example, when looking at the 256 channels of AlexNet conv5, the RMS of the difference 
between generator-produced features and bottom-up features associated with ``translate left 
by 30'' is 4.69; the mean absolute deviation is 1.042. The RMS of the difference between 
generator-produced and bottom-up associated with ``scale by 1.3x'' is 1.63; the mean 
absolute deviation is 0.46

\vspace{-2mm}
\subsection{Zero-shot learning }
\vspace{-2mm}
We have seen that the learned generators can be used to produce CNN features corresponding
to various specified transformations of provided initial CNN features. We next seek
to explore the use of these learned generators in support of a zero-shot learning task. 
We will again use our running example of rotation. 

\vspace{-1.7mm}
\paragraph{A typical example of ``zero-shot learning'': ``If it walks like
a duck...''}
\vspace{-2mm}
We first describe a typical example of zero-shot learning. Consider
the task of classifying central objects in e.g. ImageNet images of
animals. A standard zero-shot learning approach to this task involves two steps.
% One way to approach this central-object classification task
% is to regard the goal as just learning a direct mapping from raw input
% data to a vector of predicted class label probabilities. An alternative
% approach that has been explored recently breaks down this overall
% goal into two parts: 
In the first step, we learn a mapping from raw input data (for example, a picture 
of a dog) to some intermediate representation (for example, scores associated with
``semantic properties'' such as ``fur is present'', ``wings are present'', etc.). 
In the second step, we assume that we have (from Wikipedia
article text, for example) access to a mapping from the intermediate
``semantic property'' representation to class label. For example,
we expect Wikipedia article text to provide us with information such
as ``a zebra is a hoofed mammal with fur and stripes''. 

If our training data is such that we can produce accurate semantic
scores for ``hoofs are present'', ``stripes are present'', ``fur
is present'', we can potentially use the Wikipedia-derived association
between ``zebra'' and its ``semantic properties'' to bridge the
gap between the ``semantic properties'' predicted from the raw input
image and the class label associated with ``zebra''; significantly,
so long as the predicted ``semantic properties'' are accurate, the
second part of the system can output ``zebra'' whether or not the
training data ever contained a zebra. %--- this is the ``zero'' in zero-shot learning. 
% Thus, when we compose the first (learned) mapping (from raw input data to intermediate representation)
% with the second (assumed to be available) mapping (from intermediate
% representation to class label), we obtain an overall mapping that
% takes raw input data to class label. Since this two-step approach
% has the potential to produce correct labels even for classes observed
% zero times during training, it is commonly referred to as ``zero-shot
% learning''; a more accurate description might split this phrase into
% ``learning'' and ``zero-shot''. Of these two parts, the ``learning''
% comes first and is concerned with learning a mapping from raw data
% to some ``intermediate representation'' (``property scores'' in
% the animal example). By leveraging a known (from some alternative
% source) mapping from ``property scores'' to class labels, we get
% the ability to produce correct labels in ``zero-shot'' settings
% (i.e. in settings where the training data did not contain any examples
% from the class/classes that we seek to predict at test time). 
To quote the well-known aphorism: ``If it walks like a duck, swims like a
duck, and quacks like a duck, then I call that thing a duck.''

\vspace{-2mm}
\paragraph{Zero-shot learning in our context}
\vspace{-2mm}
In the typical example of zero-shot learning described above, the
task was to map raw input data to a vector of predicted class label
probabilities. This task was broken into two steps: first map the
raw input data to an intermediate representation (``semantic property
scores'', in the animal example), then map the intermediate representation
to a vector of predicted class label probabilities. The mapping from
raw input data to intermediate representation is learned during training;
the mapping from intermediate representation to class label is assumed
to be provided by background information or otherwise accessible from
an outside source (determined from Wikipedia, in the animal example).

In our setting, we have (initial image, transformed image) pairs. Our overall
goal is to determine a mapping from (initial image, transformed image) to
``characterization of specific transformation applied''. A specific
instance of the overall goal might be: when presented with, e.g., an input 
pair (image with central object, image with central object rotated 35$^\circ$) return 
output ``35$^{\circ}$''. Analogous to the animal example discussed above, we break 
this overall mapping into two steps: The first mapping step takes input pairs (initial image,
transformed image) to ``collected per-layer feature flow vector fields''. The
second mapping step takes an input of ``collected per-layer feature
flow vector fields'' to an output of ``characterization of specific
transformation applied''. Note that, in contrast to the animal example,
our context uses learning in the second mapping step rather than the
first. A specific example description of this two-part process: Take
some previously never-seen new image with a central object. Obtain
or produce another image in which the central object has been rotated
by some amount. Push the original image through AlexNet and collect
the resulting AlexNet features for all layers. Push the rotated-central-object
image through AlexNet and collect the resulting AlexNet features
for all layers. For each layer, compute the ``feature flow'' vector
field; this is the end of the first mapping step. The second mapping
step takes the collection of computed ``per-layer feature flow vector
fields'' and predicts the angle of rotation applied between the pair
of images the process started with. In our context, we use our learned generator 
in this second mapping step. We now discuss the details of our approach to
``zero-shot learning''.

\vspace{-2mm}
\paragraph{Details of our ``zero-shot learning'' task}
\vspace{-2mm}
The specific exploratory task we use to evaluate the feasibility of zero-shot
learning (mediated by top-down information distilled from the observed
behavior of network internal feature flow fields) can be described
as follows: We have generated (image-with-central-object, image-with-central-object-rotated)
pairs. We have computed feature flows for these pairs. We have performed
PCA on these feature flows to determine $U\left[\Delta\theta\right]\in\mathbb{R}^{8,522\times33}$,
an ``effective basis matrix'' associated with a rotation angle of
$\Delta\theta=10^{\circ}$. We have fit a calibration regression,
resulting in $w^{\text{lsq}}\in\mathbb{R}^{33}$, the vector of least-squares
coefficients with which we can make feature flow predictions in terms
of the ``effective basis matrix''. Our initial ``zero-shot learning'' prediction task
will be to categorize the rotation angle used in the image
pair as ``greater than 60$^\circ$'' or ``less than 60$^\circ$''.

%\begin{wrapfigure}{r}{0.3\textwidth}
\begin{figure}[!htp]
\vspace{-2mm}
\begin{centering}
\begin{tabular}{c}
\includegraphics[width=.3\textwidth]{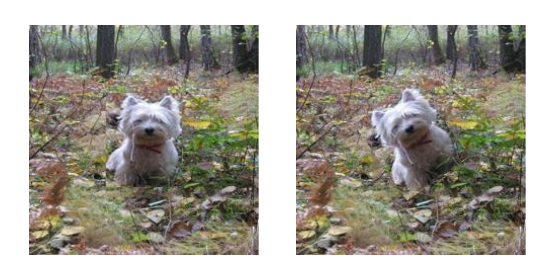}
\end{tabular}
\vspace{-2mm}
\caption{Desired categorization output: ``Rotated less than 60$^\circ$''.} %Example image pairs demonstrating the task in our investigation of ``zero-shot'' rotation angle prediction.}
\label{fig:zero_shot_pairs}
\vspace{-2mm} 
\par\end{centering}
\end{figure}
%\end{wrapfigure}

%\vspace{-1mm}
We compare the performance of our approach to a more standard approach:
We train a CNN in a ``Siamese'' configuration to take as input pairs
of the form (image, image-with-central-object-rotated) and to produce
as output a prediction of the angle of rotation between the images
in the pair. One branch of the ``Siamese'' network receives the
initial image as input; the other branch receives input with the central object 
rotated. Each branch is structured to match AlexNet layers
from conv1 up to pool5 --- that is, up to but not including fc6. We
then stack the channels from the respective pool5 layers from each branch. 
The resulting stack is provided as input to a fully-connected layer, fc6, 
with 4096 units; fc7 takes these 4096 units of input and
produces 4096 units of output; finally, fc8 produces a single scalar output --- 
probability the rotation angle used in the image
pair was ``greater than 60$^\circ$'' or ``less than 60$^\circ$''.

\vspace{-1mm}
On a test set of 1,600 previously unseen image pairs with orientation angles ranging 
through 360$^{\circ}$, our initial zero-shot learning approach yields correct categorization
74\% of the time. We structure our comparison question as ``How many image pairs are required to train the `Siamese' network
to a level of prediction performance comparable to the zero-shot approach?''
We observe that with 500 training pairs, the ``Siamese'' network
attains 62\% correct categorization; with 2,500 pairs performance improves to 64\%;
with 12,500 pairs, to 86\%; and finally with 30,000 pairs,
to 96\%.

\vspace{-2mm}
\subsection{(Network internal) ``data augmentation''}
\vspace{-2mm}
We have previously illustrated our ability to use learned generators to produce 
a variety of ``predicted'' CNN feature response maps, each of which corresponds to some 
exact CNN feature response map that would have arisen in a standard bottom-up approach; 
we will now describe how we can these learned generators to perform (network internal) 
``data augmentation''. To ground our discussion, consider an initial image $I_{j}\left[\theta_{\text{init}}\right].$
If one were to perform standard ``data augmentation'', one might
apply a variety of rotations to the initial image, say from a possible
collection of $n$ rotation angle amounts $\left\{ \left(\Delta\theta\right)_{1},\left(\Delta\theta\right)_{2},\ldots,\left(\Delta\theta\right)_{n}\right\} ,$
where we have chosen our notation to emphasize that the index is over
possible ``$\Delta\theta$'' rotation angle values. The ``data
augmentation'' process would involve $n$ corresponding images, $\left\{ I_{j}\left[\theta_{\text{init}}+\left(\Delta\theta\right)_{1}\right],\ldots,I_{j}\left[\theta_{\text{init}}+\left(\Delta\theta\right)_{n}\right]\right\} .$
Network training then proceeds in the usual fashion: for whichever
transformed input image, the corresponding AlexNet feature collection
in $\left\{ F_{j}\left[\theta_{\text{init}}+\left(\Delta\theta\right)_{1}\right],\ldots,F_{j}\left[\theta_{\text{init}}+\left(\Delta\theta\right)_{n}\right]\right\} $
would computed and used to produce loss values and backpropagate updates
to the network parameters.

Our observation is that we can use our learned generators to produce, in a 
network-internal fashion, AlexNet internal features akin to those listed in 
$\left\{ F_{j}\left[\theta_{\text{init}}+\left(\Delta\theta\right)_{1}\right],\ldots,F_{j}\left[\theta_{\text{init}}+\left(\Delta\theta\right)_{n}\right]\right\} $.
Specifically, in our running rotation example we learned to produce predictions 
(for each layer of the network) of the flow field associated with a specified rotation
angle. As mentioned previously, we refer to the learned generator at a layer 
$\ell$ associated with a rotation angle of $\Delta\theta$ as $G_{\ell}\left[\Delta\theta\right].$ 
We regard the process of applying a learned generator, for example $G_{\ell}\left[\left(\Delta\theta\right)_{1}\right]$,
to the layer $\ell$ AlexNet features, $F_{j,\ell}\left[\theta_{\text{init}}\right]$,
as a method of producing feature values akin to $F_{j,\ell}\left[\theta_{\text{init}}+\left(\Delta\theta\right)_{1}\right].$
To emphasize this notion, we will denote the values obtained
by applying the learned generator flow field $G_{\ell}\left[\left(\Delta\theta\right)_{1}\right]$
to the layer $\ell$ AlexNet features, $F_{j,\ell}\left[\theta_{\text{init}}\right]$
as $\Phi_{j,\ell}\left[\theta_{\text{init}};\left(\Delta\theta\right)_{1}\right]$
(with the entire collection of layers denoted as $\Phi_{j}\left[\theta_{\text{init}};\left(\Delta\theta\right)_{1}\right]$).
Using our newly-established notation, we can express our proposed
(network internal) ``data augmentation'' as follows: From some initial
image $I_{j}\left[\theta_{\text{init}}\right],$ compute AlexNet
features $F_{j}\left[\theta_{\text{init}}\right].$ For any desired
rotation, say $\Delta\theta$, determine the associated learned generator flow field
$G_{\ell}\left[\Delta\theta\right].$ Apply this generator flow field to, for
example, $F_{j,\ell}\left[\theta_{\text{init}}\right]$ to obtain
``predicted feature values'' $\Phi_{j,\ell}\left[\theta_{\text{init}};\Delta\theta\right].$
The standard feedforward computation can then proceed from layer $\ell$
to produce a prediction, receive a loss, and begin the backpropagation
process by which we can update our network parameters according to
this (network internal) ``generated feature example''. 

\begin{table}[!htp]
\vspace{-2mm}
{\footnotesize
\begin{center}
\caption{\label{tab:imagenet} ImageNet validation set accuracy (in \%).}
\vspace{-3mm}
\begin{tabular}{lcccc} 
\multicolumn{1}{l}{Method}                            &\shortstack{top-1 \\ m-view} & \shortstack{top-5 \\ m-view}
\\ \hline 
\scriptsize{AlexNet \citep{krizhevsky2012imagenet} }                       & 60.15   & 83.93 \\
\scriptsize{AlexNet after 5 additional epochs of generator training }     & 60.52   & 84.35 \\
\hline 
\end{tabular}
\end{center}
}
\vspace{-2mm}
\end{table}

\begin{figure*}[!htp]
\vspace{-3mm}
\begin{center}
\begin{tabular}{c}
\includegraphics[width=.6\textwidth]{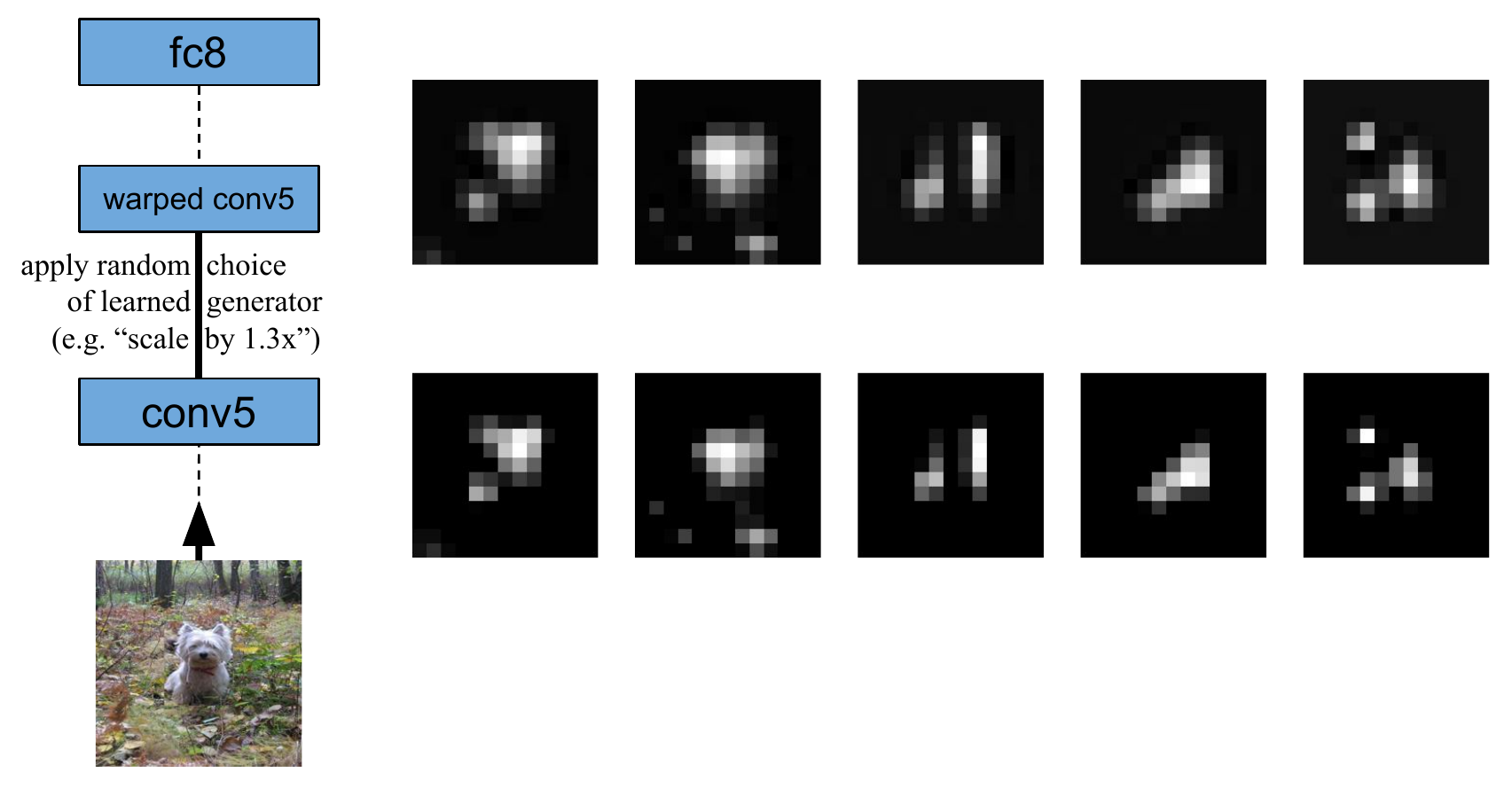}
\end{tabular}
\end{center}
\vspace{-6mm}
\caption{\footnotesize Illustration of network internal data augmentation, more succinctly described as ``generator training''. 
On the left, we show a schematic of the modified AlexNet architecture we use. 
The primary difference is the incorporation at conv5 of a module applying a randomly 
selected learned generator flow field. On the right, we provide comparison between 
five selected conv5 channels: (lower row) before applying the ``scale 1.3x'' learned generator flow field; (upper row) after applying the generator flow field.}
\label{fig:network_internal_data_augmentation}
\vspace{-2mm}
\end{figure*}

The backpropagation process involves a subtlety. Our use of the generator to means 
that forward path through the network experiences a warp in the features. To correctly 
propagate gradients during the backpropagation process, the path that the gradient 
values follow should experience the ``(additive) inverse'' of the forward warp. 
We can describe the additive inverse
of our gridded vector field fairly simply: Every vector in the initial
field should have a corresponding vector in the inverse field; the
component values should be negated and the root of the ``inverse
vector'' should be placed at the head of the ``forward vector''.
The ``inverse field'' thus cancels out the ``forward field''.
Unfortunately, the exact ``inverse vector'' root locations will
not lie on the grid used by the forward vector field. We obtain an
approximate inverse vector field by negating the forward vector field components. 
In tests, we find that this approximation is often quite good; see Fig. \ref{fig:conv1_apply_generator_and_negated_generator}. Using this approximate inverse warp, our learned generator 
warp can be used in the context of network internal data augmentation
during training. See the Fig. \ref{fig:network_internal_data_augmentation} 
for an illustration.

We now discuss the use of our proposed network internal data augmentation
to improve the performance of AlexNet on ImageNet. We train using
the 1.3M images of the ILSVRC2014 CLS-LOC task. For each batch of
256 images during training, we randomly select one of six of our learned
generator flow fields to apply to the initial features in conv5. Specifically,
we randomly select from one of +30$^{\circ}$ rotation, -30$^{\circ}$
rotation, 1.3x scaling, 0.75x scaling, translation 30 pixels left, or
translation 30 pixels up. We apply the (approximate) inverse warp when
backpropagating through conv5. We evaluate performance on the 50k
images of ILSVRC2014 CLS-LOC validation set; see Table \ref{tab:imagenet}.

\vspace{-2mm}
\paragraph{Acknowledgments} This work is supported by NSF IIS-1216528 (IIS-1360566), 
NSF award IIS-0844566 (IIS-1360568), and a Northrop Grumman Contextual Robotics
grant. We gratefully acknowledge the support of NVIDIA Corporation with the donation 
of the Tesla K40 GPU used for this research. We thank Chen-Yu Lee and Jameson Merkow 
for their assistance, and Saining Xie and Xun Huang for helpful discussion.
\vspace{-2mm}

\bibliography{merged}
\bibliographystyle{iclr2016_conference}

\clearpage
\setcounter{section}{0}
\renewcommand{\thesection}{A\arabic{section}}
\setcounter{table}{0}
\renewcommand{\thetable}{A\arabic{table}}
\setcounter{figure}{0}
\renewcommand{\thefigure}{A\arabic{figure}}

\section{Supplementary Materials}

% \begin{figure*}[!htp]
% \vspace{-3mm}
% \begin{center}
% \begin{tabular}{c}
% % \includegraphics[height=.8\linewidth]{figs/network_internal_data_augmentation_with_two_different_warps_and_backprop_arrows.pdf}
% \includegraphics[height=.8\linewidth]{figs/network_internal_data_augmentation_with_feature_visualization.pdf}
% \end{tabular}
% \end{center}
% \vspace{-6mm}
% \caption{\footnotesize Illustration of network internal data augmentation, more succinctly described as ``generator training''. 
% On the left, we show a schematic of the modified AlexNet architecture we use. 
% The primary difference is the incorporation at conv5 of a module applying a randomly 
% selected learned generator flow field. On the right, we provide comparison between 
% five selected conv5 channels: (lower row) before applying the ``scale 1.3x'' learned generator flow field; (upper row) after applying the generator flow field.}
% \label{fig:network_internal_data_augmentation}
% \vspace{-2mm}
% \end{figure*}

\begin{figure*}[!htp]
\vspace{-3mm}
\begin{center}
\begin{tabular}{c}
\includegraphics[width=0.6\textwidth]{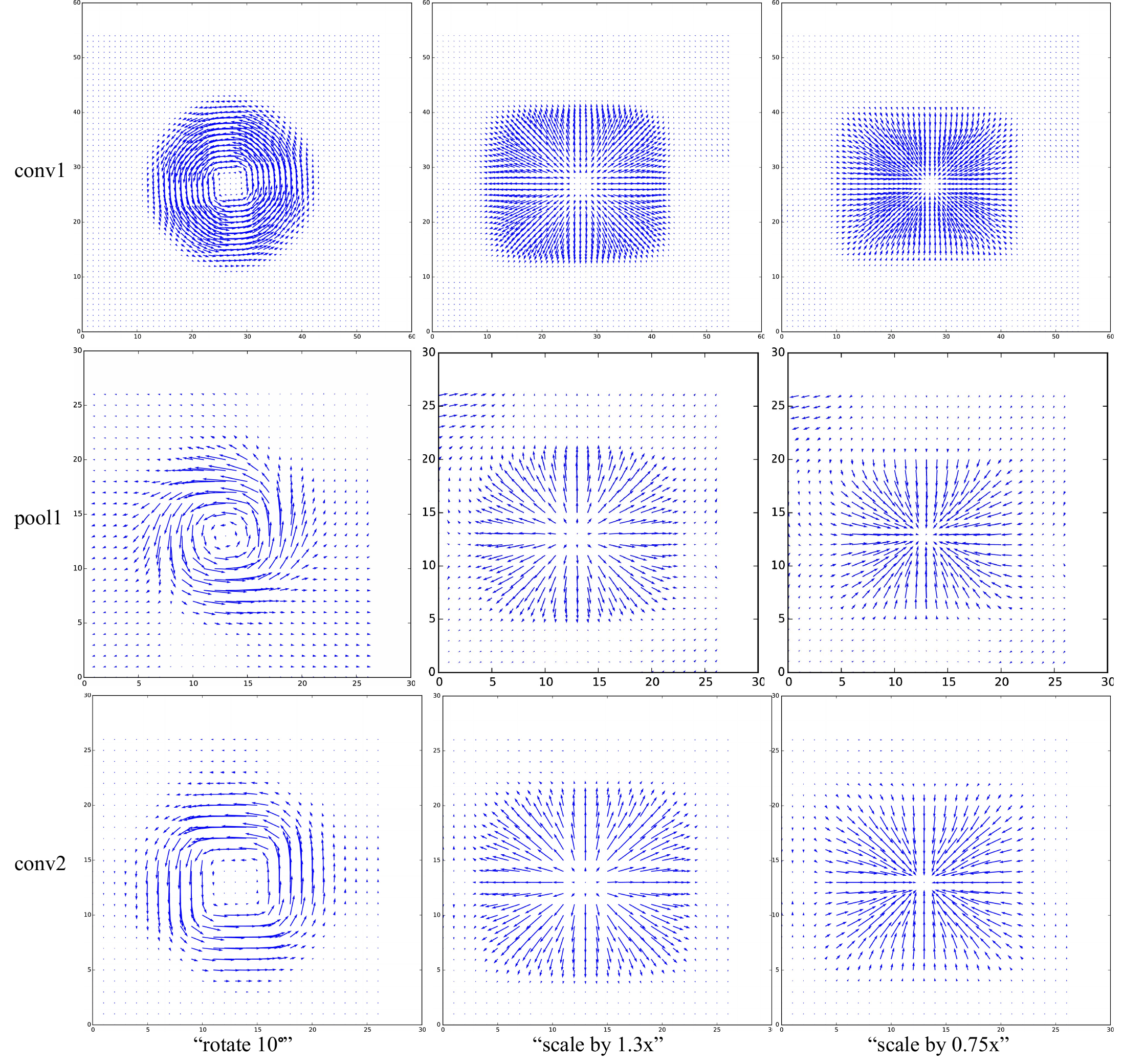}
\end{tabular}
\end{center}
\vspace{-6mm}
\caption{\footnotesize Visualizations of the mean ``feature flow'' as computed across the 91 ``original images'' whose 
selection is described in Section \ref{sec:image_selection}. The leftmost column contains visualizations computed 
from features arrived at with input image pairs that differ by 10$^\circ$ rotation of the central object; the center column, 
with input image pairs that differ by the central object being scaled by 1.3x; the rightmost column, with input image pairs 
that differ by the central object being scaled by 0.75x. Moving from top to bottom within each column, the feature 
flow fields are shown, respectively, for conv1, then pool1, then conv2.}
\label{fig:feature_flow_fields_conv1_pool1_conv2}
\vspace{-2mm}
\end{figure*}

\begin{figure*}[!htp]
\vspace{-3mm}
\begin{center}
\begin{tabular}{c}
\includegraphics[width=.6\textwidth]{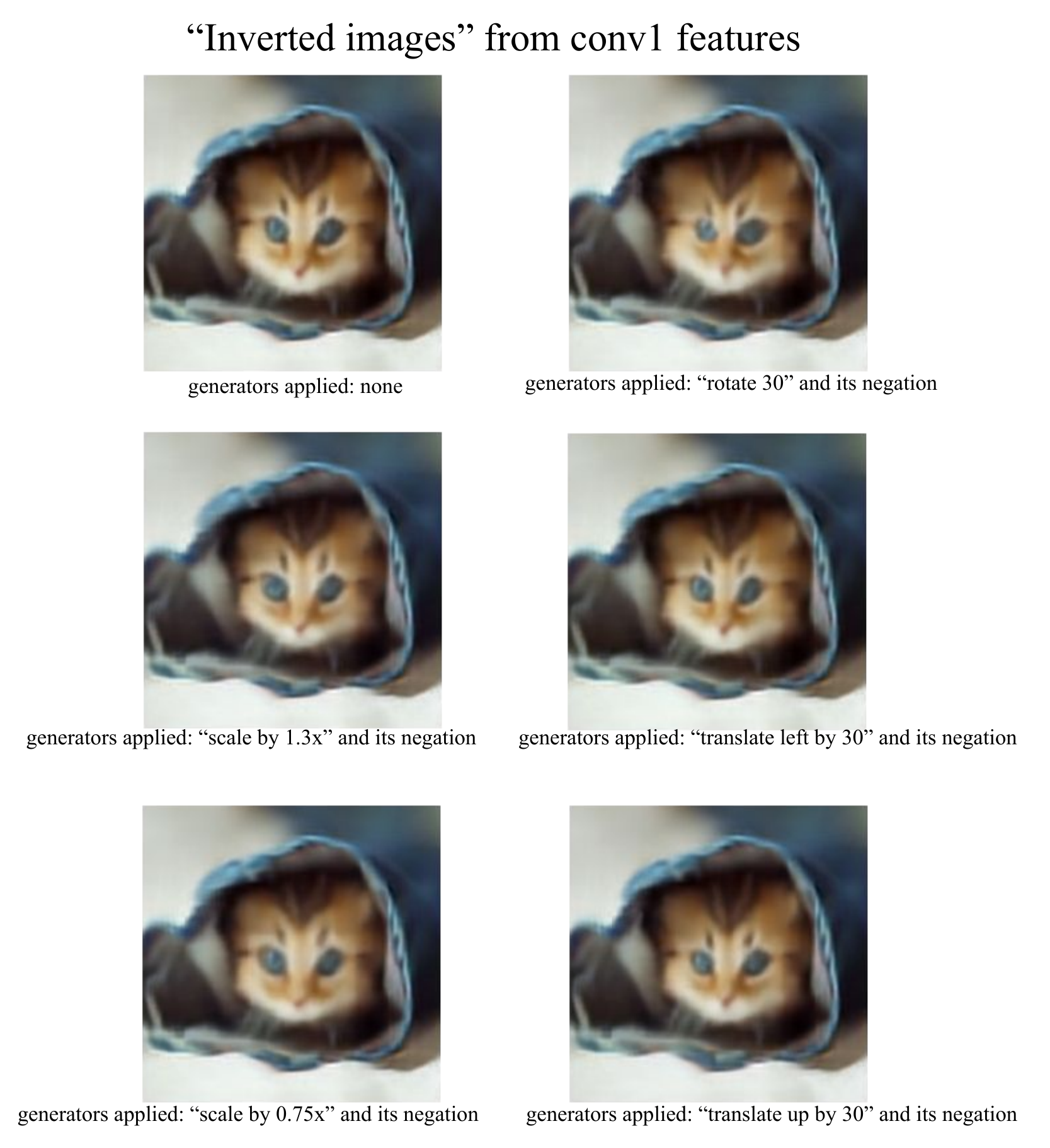}
\end{tabular}
\end{center}
\vspace{-6mm}
\caption{\footnotesize Here we confirm that the ``negation'' of a generator flow field is (both qualitatively and quantitatively) 
a good approximation to the additive inverse of that generator flow field. Since the inverted images from conv1 have more detail, we 
perform our qualitative evaluation with conv1. Since our actual training uses conv5, we perform our quantitative 
evaluation with conv5. In each entry above, we begin with AlexNet conv1 features from the 
original ``taco cat'' image. The ``inverted image'' \citep{dosovitskiy2015inverting} corresponding to these untouched features 
is found in the top left. In each other entry we apply a different learned generator followed by its ``negation''. The 
close correspondence between the images ``inverted'' from the resulting features and the image ``inverted'' from the 
untouched features confirms the quality of the approximation. Moving on to quantitative evaluation, we find that the 
feature values arising from applying a generator flow field followed by its negation differs from the original AlexNet 
conv5 feature values across the 256 channels of conv5 as follows: approximate inverse of ``rotate -30'' yields 0.37 RMS 
(0.09 mean absolute difference); approximate inverse of ``scale by 1.3x'' yields 0.86 RMS (0.19 mean absolute difference); for 
``translation 30 left'', the approximation incurs error at the boundaries of the flow region, yielding 3.96 RMS (but 0.9 mean 
absolute deviation). }
\label{fig:conv1_apply_generator_and_negated_generator}
\vspace{-2mm}
\end{figure*}

% \begin{figure*}[!htp]
% \vspace{-3mm}
% \begin{center}
% \begin{tabular}{c}
% % \includegraphics[height=.8\linewidth]{figs/network_internal_data_augmentation_with_two_different_warps_and_backprop_arrows.pdf}
% \includegraphics[height=.7\linewidth]{figs/conv2_apply_generator_and_negated_generator.pdf}
% \end{tabular}
% \end{center}
% \vspace{-6mm}
% \caption{\footnotesize Further ``inverted images'' illustrating the quality of the 
% ``negation'' as an approximate additive inverse of a learned generator flow field; here with AlexNet conv2 features.}
% \label{fig:conv2_apply_generator_and_negated_generator}
% \vspace{-2mm}
% \end{figure*}

\end{document}